\documentclass[10pt,twocolumn,letterpaper]{article}

\usepackage{wacv}
\usepackage{times}
\usepackage{epsfig}
\usepackage{graphicx}
\usepackage{amsmath}
\usepackage{amssymb}
\usepackage{multirow}
\usepackage{booktabs}
\DeclareMathOperator{\arctantwo}{arctan2}

\def\wacvPaperID{1642} % Enter the WACV Paper ID here

\wacvfinalcopy % *** Uncomment this line for the final submission

\ifwacvfinal
\def\assignedStartPage{1} % *** Enter the assigned starting page number (instead of 9876)
\fi

\ifwacvfinal
\usepackage[breaklinks=true,bookmarks=false]{hyperref}
\else
\usepackage[pagebackref=true,breaklinks=true,colorlinks,bookmarks=false]{hyperref}
\fi

\ifwacvfinal
\else
\fi

\begin{document}

%%%%%%%%% TITLE
\title{Weakly-Supervised Optical Flow Estimation for Time-of-Flight}

\author{Michael Schelling, Pedro Hermosilla, Timo Ropinski\\
Ulm University, Germany\\
{\tt\small https://github.com/schellmi42/WFlowToF}
% \and
}

\maketitle

\begin{figure*}[t!]
    \centering
    \includegraphics[width =\linewidth]{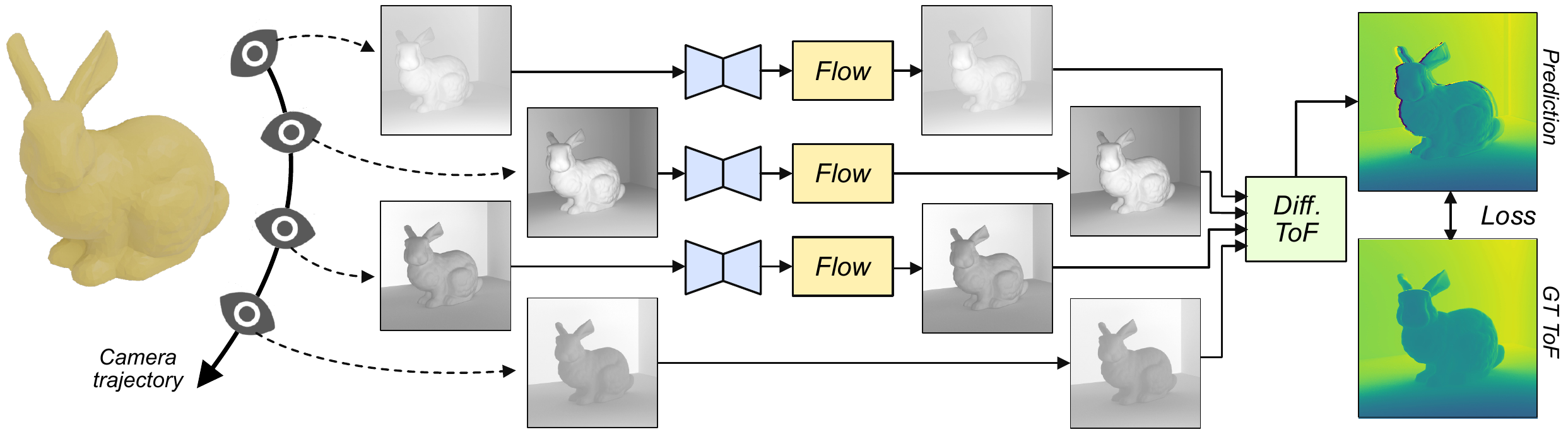}
    \caption{Illustration of the flow estimation. 
    Given iToF measurements at subsequent time steps, a network is used to predict optical flows, in order to align the images to the reference image (bottom row).
    From the warped measurements a ToF depth image can be reconstructed.
    We propose to supervise the training directly on this ToF depth, and propagate gradients through the ToF depth computation.
    This figure shows the single frequency, single-tap case with four measurements.
    Note the modality change in the input due to different phase shifts $\theta$.}
    \label{fig:teaser}
\end{figure*}
\thispagestyle{empty}

%%%%%%%%% ABSTRACT  

% 2000 characters maximum!

% Indirect Time-of-Flight (iToF) cameras are a widespread type of 3D sensor.
% However, due to the indirect measure ment principle, multiple captures are needed, which intro duces misalignments, if the scene is not static.
% While recent approaches to correct iToF depths achieve high correction capabilities, motion artifacts are mostly treated as an additional error in the reconstructed depth images, without special considerations. 
% In this work we propose a training algorithm for Optical Flow (OF) prediction networks, which enables the network to resolve the multi-view, multi-modality OF estimation in an unsupervised fashion and finally to reduce motion artifacts in iToF depth images.

\begin{abstract}
Indirect Time-of-Flight (iToF) cameras are a widespread type of 3D sensor, which perform multiple captures to obtain depth values of the captured scene. While recent approaches to correct iToF depths achieve high performance when removing multi-path-interference and sensor noise, little research has been done to tackle motion artifacts. In this work we propose a training algorithm, which allows to supervise Optical Flow (OF) networks directly on the reconstructed depth, without the need of having ground truth flows. We demonstrate that this approach enables the training of OF networks to align raw iToF measurements and compensate motion artifacts in the iToF depth images. The approach is evaluated for both single- and multi-frequency sensors as well as multi-tap sensors, and is able to outperform other motion compensation techniques.
\end{abstract}

%%%%%%%%% BODY TEXT
%-------------------------------------------------------------------------
\section{Introduction}
Time-of-Flight (ToF) cameras are sensors that aim to capture depth images by measuring the time the light needs to travel from a light source on the camera to an object and back to the camera sensor. Apart from direct ToF cameras, such as LiDAR, which register the time of incoming reflections of a light pulse at a high temporal resolution, another common and cost-efficient approach are indirect ToF (iToF) cameras, which do not require as precise measuring devices.
One realization of iToF devices are Amplitude Modulated Continuous Wave (AMCW) ToF sensors, as for example used in the Kinect system. These sensors continuously illuminate the scene with a periodically modulated light signal and aim to retrieve the phase offset between the emitted and the retrieved signal, which gives information about the travel time of the signal~\cite{hansard2012tof_principles}.
In order to retrieve the phase offset it is necessary to perform multiple captures, which makes this approach sensible to movements of both, the camera and the objects in the illuminated scene.
As the measurements are taken with differing sensor settings, so called multi modality, standard Optical Flow (OF) algorithms achieve only low performance, and hence require adaptation.
While there are works that investigate the compensation of motion using OF, they are only applicable to specific sensor types~\cite{lindner2009compensation, hoegg2013real} or require carefully designed datasets~\cite{guo2018tackling} to train OF networks.
Hence, it is still a common approach to merely detect motion artifacts and mask the affected pixels in the final depth image, as is for example realized by the LF2-algorithm~\cite{lingzhu2016LF2} for the Kinect sensor.

In this work, we propose a training algorithm for OF networks which allows to supervise the flow prediction using the ToF depth image, without the need to directly supervise the predicted flow, see Fig.~\ref{fig:teaser}.
To this end, we analyze the ToF depth computation to provide reliable and stable gradients during training.
Further, we introduce a set of regulatory losses, which guide the network towards predicting flows, that are consistent with the underlying images.

%-------------------------------------------------------------------------
\section{Technical Background}
In this section, we briefly describe iToF cameras.

\paragraph{ToF Working Principle.}
An AMCW iToF camera emits a modulated light signal $s(t)$, which is correlated at the sensor with a phase shifted version of the emitted signal $s(t+\theta)$ during the exposure time.
The resulting measurement $m$ is repeated sequentially for different phase shifts $\theta$, from which the distance $d$ is retrieved indirectly by estimating the phase shift $\Delta\varphi$ of the signal $s$ when arriving at the sensor. 
In the common case of four measurements $m_0,\dots,m_3$ at $\theta\in\{0, \pi/2, \pi, 3\pi/2\}$, the distance $d$ is retrieved as
\begin{align}
    \Delta\varphi &= \arctan\left( \frac{m_3 - m_1}{m_0 - m_2} \right), \label{eq:tof_phase}\\
    d_{ToF} &= \frac{c\cdot \Delta\varphi}{4\pi f}, \label{eq:tof_depth}
\end{align}
where $c$ is the speed of light, and $f$ is the modulation frequency of the signal $s$~\cite{hansard2012tof_principles}. Due to the periodic nature of Eq.~\eqref{eq:tof_phase}, the reconstructed $d_{ToF}$ is only unambiguous up to a maximum distance of 
\begin{align}
    d_{max} = c / (2f),
\end{align}
specifically, $d_{ToF} = d\mod d_{max}$, where the distance $d$ is referred to as \emph{depth}, as is common practice in the area of ToF imaging. The so called phase wrapping of $d_{ToF}$ is typically resolved by using additional measurements at different frequencies $f$~\cite{hansard2012tof_principles}.

However Eq.~\eqref{eq:tof_depth} is based on the assumptions that, (a) only the direct reflection $s(t + \Delta\varphi)$ is captured and (b) the scene is static between the different captures. 
While (a) has been dealt with to a large extent in recent work on correcting iToF depths~\cite{agresti2018deep, marco2017deeptof, schelling2022radu}, only little research has been done to reduce motion artifacts stemming from (b).

%-------------------------------------------------------------------------
\paragraph{Multi-Tap Sensors.}
A realization of iToF sensors are so called \emph{multi-tap} sensors, which are able to capture multiple measurements of $m_\theta$ in parallel.
The most widespread approach are two-tap sensors, which allow the capture of $m_{A,i} = m_i$ and $m_{B,i} = m_{i+2}$ at the same time, by sorting the electrons generated by incoming photons into two quantum wells using a modulated electric field~\cite{schmidt2011analysis}.
Internally, these two measurements are used to compensate for hardware inaccuracies and reduce noise ~\cite{hoegg2013real} by computing: 
\begin{align}
    m_i &= m_{A, i} - m_{B, i}.
\end{align}
In order to make direct use of $m_A, m_B$ in Eq.~\eqref{eq:tof_phase}, it is necessary to calibrate the differences in the photo responses~\cite{schmidt2011analysis}
\begin{align}
    m_{A,i} & = r_\theta(m_{B,i+2}),
\end{align}
which doubles the effective frame rate, and reduce, but not eliminate, motion-artifacts.
% it is necessary to fit an intensity correction function $f$ ~\cite{lindner2009compensation} by minimizing
% \begin{align}
%     \sum_{\theta}\big(f(m_{A,\theta}) + f(m_{B,\theta})\big) = I_{ref},
% \end{align}
% with a reference Intensity $I_ref$.
Recently also prototypes for four-tap sensors have been developed~\cite{chen2022fourtap, keel2019fourtap}, which in the future might eliminate motion artifacts in single-frequency captures, but not in multi-frequency sensors.

%-------------------------------------------------------------------------
\section{Related Work}
This section briefly summarizes previous work on related fields.

%-------------------------------------------------------------------------
\paragraph{ToF Motion Artifact Correction.}
% (PMD: two-tap sensor: 0,180, and 90, 270 are recorded simultaneously)\\
% Lottner \etal 2007, not learned, uses bilateral filtering, not really well developed
Early methods on motion compensation used detect-and-repair approaches~\cite{schmidt2011analysis, hansard2012tof_principles}, \eg by performing bilateral filtering~\cite{lottner2007movement}.
One of the first methods to resolve movement artifacts using optical flow was introduced by Lindner \etal~\cite{lindner2009compensation} who aim to tackle the cross modality through a correction scheme to compute intensity images from two-tap captures, which can be used as input to a standard OF algorithm. 
Based on this method, Hoegg \etal ~\cite{hoegg2013real} derived optimizations for the OF prediction algorithm by incorporating motion detection and refining the spatial consistency to achieve real-time performance.
The performance of these approaches was further improved with the calibration of Gottfried \etal~\cite{gottfried2014time}.
In contrast we integrate the entire computational flow, from raw iToF measurement to depth reconstruction into our optimization pipeline.

The first learned approach was presented by Guo \etal~\cite{guo2018tackling}, who provide methods to correct errors for the Kinect2 sensor, including an encoder-decoder network for OF prediction.
To enable the supervised learning of motion compensation, a specific dataset is generated, which allows for simulating linear movements in the image domain, while separating the motion of foreground and background.
Contrarily, we propose a weakly supervised training, which does not require flow labels, and instead uses ToF depths for supervision, which are available in existing iToF datasets.

%-------------------------------------------------------------------------
\paragraph{Optical Flow.}
Recent works on OF regression rely on neural networks, which have proven to outperform traditional approaches~\cite{sun2018pwc}. 
The typical design, using shared image encoders and a latent cost volume, was first introduced Dosovitskiy \etal~\cite{dosovitskiy2015flownet} in their FlowNetC architecture, alongside the FlowNetS network, which uses a encoder decoder architecture. 
Subsequent, a large literature on various applications~\cite{zheng2020optical, li2019rainflow} and formulations~\cite{aleotti2021learning, janai2018unsupervised} in the field of motion estimation emerged.
In order to reduce the computational costs, Sun \etal~\cite{sun2018pwc} introduced a hierarchical architecture with coarse-to-fine warping in their Pyramid-Warping-Cost-volume (PWC)  network.
This design was further refined by Kong \etal~\cite{kong2021fastflownet} in their FastFlowNet (FFN) architecture, which reduced the computational complexity and achieves fast inference times.

To overcome the need of generating ground truth flows for a supervised training, unsupervised approaches~\cite{jonschkowski2020matters, ren2017unsupervised, yu2016back, janai2018unsupervised} optimize the photometric consistency between images and apply regularizations to refine the flow prediction.

% Jonschkowski \etal 2020, analysis of unsupervised optical flow networks, various improvements.

% Sun \etal 2018, PWC-Net, probably SotA, together with RAFT, more lightweight than FlowNet2.

% Teed and Deng 2020, RAFT, Iterative approach using a recurrent unit, current SotA, quite expensive(?).

% Kong \etal 2021, FastFlowNet, lightweight approach for fast execution.

% Zheng \etal 2020, optical flow for underexposed images (high noise levels).

% Aleotti \etal 2021, optical flow from still images.

%-------------------------------------------------------------------------
\paragraph{ToF Correction.}
The occurrence of Multi-Path-Interference (MPI) is the main source of errors in iToF depth reconstructions.
Consequently, existing works on correcting iToF data focus on removing MPI artifacts.
As with OF prediction, 2D neural networks have proven to achieve high noise removal performance~\cite{agresti2018deep, marco2017deeptof, su2018deep, guo2018tackling, dong2020spatial}.
However, also other learned approaches have been investigated recently, such as reconstructing the transient response~\cite{buratto2021deep, gutierrez2021itof2dtof} or using 3D point networks~\cite{schelling2022radu}.

% \begin{figure*}
% \begin{center}
% \fbox{\rule{0pt}{2in} \rule{.9\linewidth}{0pt}}
% \end{center}
%    \caption{Example of a short caption, which should be centered.}
% \label{fig:short}
% \end{figure*}

%------------------------------------------------------------------------
% \section{Problem Statement}

%------------------------------------------------------------------------
\section{Method}\label{sec:method}

\begin{figure*}[t!]
    \centering
    \includegraphics[width =\linewidth]{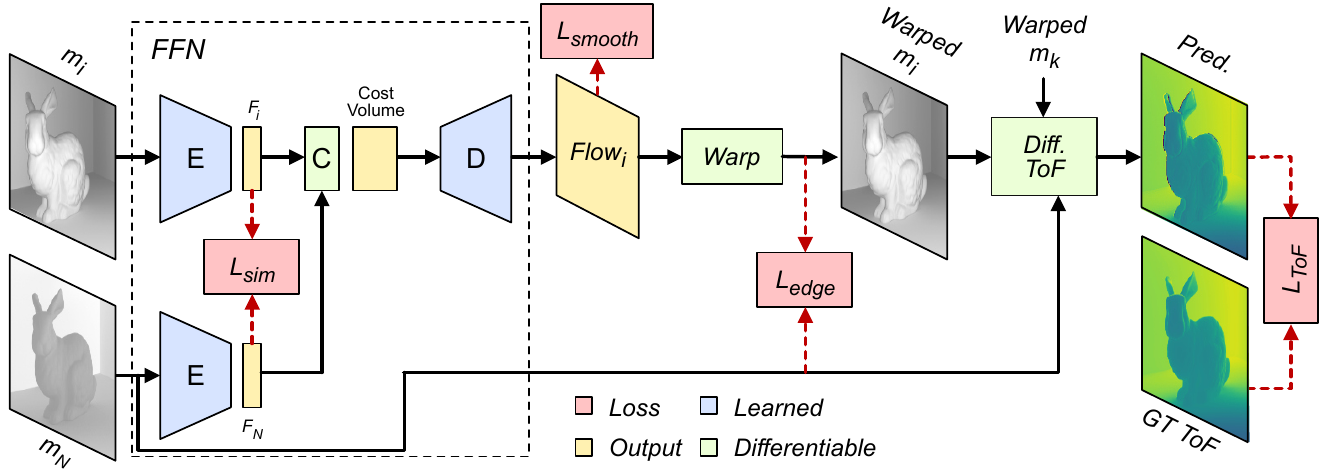}
    \caption{Overview over the loss functions used in this work.
    Our main loss is the ToF-loss $\mathcal{L}_{ToF}$ (right), which is computed on the reconstructed ToF depth using a differentiable operation, and is adapted to provide phase unwrapped gradients.
    To constrain the flow prediction the loss $\mathcal{L}_{smooth}$ (top) is used to regularize the flow, and an additional regularization on the warped image $m_i$ is given through the loss $\mathcal{L}_{edge}$ (center). 
    Finally, the loss $\mathcal{L}_{sim}$ aims to create consistency between the latent representations inside the network.
    Note: The losses $\mathcal{L}_{ToF}$ and $\mathcal{L}_{sim}$ are computed over all $i$.
    This figure shows the single-tap case, where only one measurement is taken per time step.}
    
    \label{fig:losses}
\end{figure*}
    
    In this work we propose a weak supervision of an OF network using the ToF-depth $d_{ToF}$ as label, without providing ground truth flow vector fields.
    In order to enable training using depth labels, the phase wrapping discontinuities in Eq.~\eqref{eq:tof_phase} of the $\arctan$ function require consideration, and regularizations on the flow prediction need to be established to predict consistent flows without direct supervision.
    % This introduces two main challenges (a) the phase wrapping discontinuities in Eq.~\eqref{eq:tof_phase} of the $\arctan$ function and (b) the not well-defined problem statement of the under-determined problem of aligning the four measurements $m_\theta$ using a single valued optimization function.
    
    We consider an OF network $g:(\{m_i\}_{i=0}^{N-1},m_N)\rightarrow \{V_i\}_{i=0}^{N-1}$, which predicts a set of optical flows $V_i$ for a set of measurements $m_i$, in order to align them to a measurement $m_N$ taken at the reference time step.
    The standard photometric loss in this setting would be given as
    \begin{align}
        \hat{m}_i &= warp(m_i, V_i)  \\
        \mathcal{L}_{photo} &= \sum_i\| \hat{m}_i- m_{i}^{GT}\|_1 \label{eq:L_photo},
    \end{align}
    where $m_i^{GT}$ is taken at the same time step as $m_N$.
    
    Instead, we propose to supervise the network $g$ indirectly on the reconstructed depth using the ToF depth $d_{ToF}$ without motion as target.
    To increase the numerical stability we formulate the reconstructed depth $\hat{d}$ as 
    
    \begin{align}
        s &= \text{sign}(\hat{m}_0 - \hat{m}_2) \\
        \hat{d} &= \frac{c}{4\pi f}\arctan\left( \frac{\hat{m}_3 - \hat{m}_1}{\hat{m}_0 - \hat{m}_2 + s\cdot\epsilon} \right), \label{eq:tof_phase_stable} \\
        \mathcal{L}_{ToF} &= \|\hat{d} - d_{ToF}\|_1 \label{eq:L_tof},
    \end{align}
    which avoids singularities as the denominator in Eq.~\eqref{eq:tof_phase_stable} is strictly positive for $\epsilon>0$.
    The implementation of Eq.~\ref{eq:L_tof} on commonly used learning packages with auto-differentiable features, such as Pytorch~\cite{NEURIPS2019_9015} or JAX~\cite{jax2018github}, allows to train the flow network $g$ in a weakly-supervised fashion.

\subsection{Phase Unwrapping}
    
    The phase wrapping in the above formulation can be tackled by generating multiple candidate depths $\hat{d}_k = \hat{d} + k\cdot d_{max}$ and using the one closest to the label as prediction
    
    \begin{align}
        \hat{d}_k &= \hat{d} + k\cdot d_{max} \\
        \mathcal{L}_{ToF,PU} &= \min\{\|\hat{d}_k - d_{ToF}\|_1\ \big|\ k\in\mathbb{Z}\}.\label{eq:candidate_depths}
    \end{align}
    As both $\hat{d}$ and $d_{ToF}$ are in the range of $[0,d_{max})$, the candidate space is reduced to $k\in\{-1,0,1\}$ and the minimization in Eq.~\eqref{eq:candidate_depths} can be realized by a simple lookup table
    
    \begin{align}
        \hat{d}-d_{ToF}&\in(-d_{max}, d_{max}/2]: &&k=-1,\\ 
        \hat{d}-d_{ToF}&\in(-d_{max}/2, d_{max}/2]: &&k=0,\\ 
        \hat{d}-d_{ToF}&\in(d_{max}/2, d_{max}): &&k=1.
    \end{align}
    However, during training only the gradients of $\mathcal{L}_{ToF,PU}$ are relevant, which can be derived from the lookup table as
    
    \begin{align}
         \nabla\mathcal{L}_{ToF,PU} = 
     \left\{
     \begin{aligned}
    	\nabla \mathcal{L}_{ToF}, &\qquad 0\leq \mathcal{L}_{ToF} < {d_{max} / 2},\\
    	-\nabla \mathcal{L}_{ToF}, &\qquad \mathcal{L}_{ToF} \geq {d_{max} / 2},\\
    	\end{aligned}
    	\right.\label{eq:grads_PU}
    \end{align}
    and can thus be directly computed from Eq.~\eqref{eq:L_tof}.
    This allows a computational cheap and elegant implementation of the phase unwrapping, by only adjusting the gradients of $\mathcal{L}_{ToF}$, Eq.~\eqref{eq:L_tof}, based on the conditions in Eq.~\eqref{eq:grads_PU} in the backpropagation step during the training of $g$.

\subsection{Regularization}
    By regularizing the predictions, additional constraints for the predicted flows $V_i$ are established, which enables the network to produce coherent predictions without using flow labels.
    We use two additional regularization losses, a smoothing loss $\mathcal{L}_{smooth}$ and an edge-aware loss $\mathcal{L}_{edge}$.
    
    For smoothing we adapt the formulation of Jonschkowski \etal~\cite{jonschkowski2020matters} to our setting
    \begin{align}
        \mathcal{L}_{smooth} = \sum_{i,j} %\sum_{n=1}^N \sum_{j=1}^M \sum_{i=1,2}
        \exp\left(-\lambda\left|\frac{\partial m_i}{\partial x_j}\right|\right) \cdot \left|\frac{\partial V_i}{\partial x_j}\right| \label{eq:L_smooth},
    \end{align}
    where $\lambda$ is an edge weighting factor and $x_0, x_1$ are the two image dimensions. 
    This loss penalizes high gradients on $V_i$ in homogeneous regions of $m_i$, \ie regions where $m_i$ has small gradients.
    The intuition of $\mathcal{L}_{smooth}$ is that homogeneous regions are expected to move in the same direction.
    
    To further regularize the network to predict correctly aligned object boundaries, we introduce an edge-aware loss
    % \begin{align}
    %     \mathcal{L}_{edge} = \frac{1}{N} \sum_{n,i}%\sum_{n=1}^N \sum_{f=1}^F\sum_{i=1,2}
    %     \exp\left(-\left(\epsilon+\left|\frac{\partial I}{\partial x_i}\right|\right)^{-1}\right) \cdot \left(\left|\frac{\partial \hat{I}}{\partial x_i}\right| + s\right)^{-1} \label{eq:L_edge}
    % \end{align}
    \begin{align}
        \mathcal{L}_{edge} = \sum_{i,j}%\sum_{n=1}^N \sum_{f=1}^F\sum_{i=1,2}
        \exp\left(\frac{-1}{\epsilon+\left|\frac{\partial m_N}{\partial x_j}\right|}\right) \cdot \frac{1}{\left|\frac{\partial \hat{m}_i}{\partial x_j}\right| + s} \label{eq:L_edge_short},
    \end{align}
    where $\epsilon$ is a small constant for numerical stability and the shift $s$ is used to provide an upper bound on the gradients of $\mathcal{L}_{edge}$.
    This loss penalizes small gradients in the warped measurements $\hat{m}_i$ in regions where $m_N$ has large gradients, \ie regions where $m_N$ has edges.
    The intuition of $\mathcal{L}_{edge}$ is that boundaries of objects can be expected to create edges in the measurements independent of the modality.
    
    Note that $\mathcal{L}_{smooth}$ acts on the flows $V_i$ whereas $\mathcal{L}_{edge}$ is computed on the warped measurements $\hat{m}_i$, see Fig.~\ref{fig:losses}.

\subsection{Cross Modality}
    To guide the network towards learning latent representations $F_i$, see Fig.~\ref{fig:losses}, that are robust to the input modality, we make use of a latent similarity loss on the column vectors $F_i(k,l)$ of the latent representation in $g(m_i)$, inspired by the formulation of contrastive learning 
    
    \begin{align}
        \mathcal{L}_{sim} = \sum_{i\neq j}\sum_{k,l}L\bigg(F_i(k, l), F_i(k,l)\bigg), \label{eq:L_sim}
    \end{align}
    where $L$ is a similarity loss, \eg L$_1$, L$_2$, the cosine-similarity or a cost function.
    
    During training we optimize the similarity loss on static scenes, without motion.
    An overview of all losses and their integration in the computational flow are shown in Fig.~\ref{fig:losses}.

\begin{figure*}[t]
    \centering
    \includegraphics[width =\linewidth]{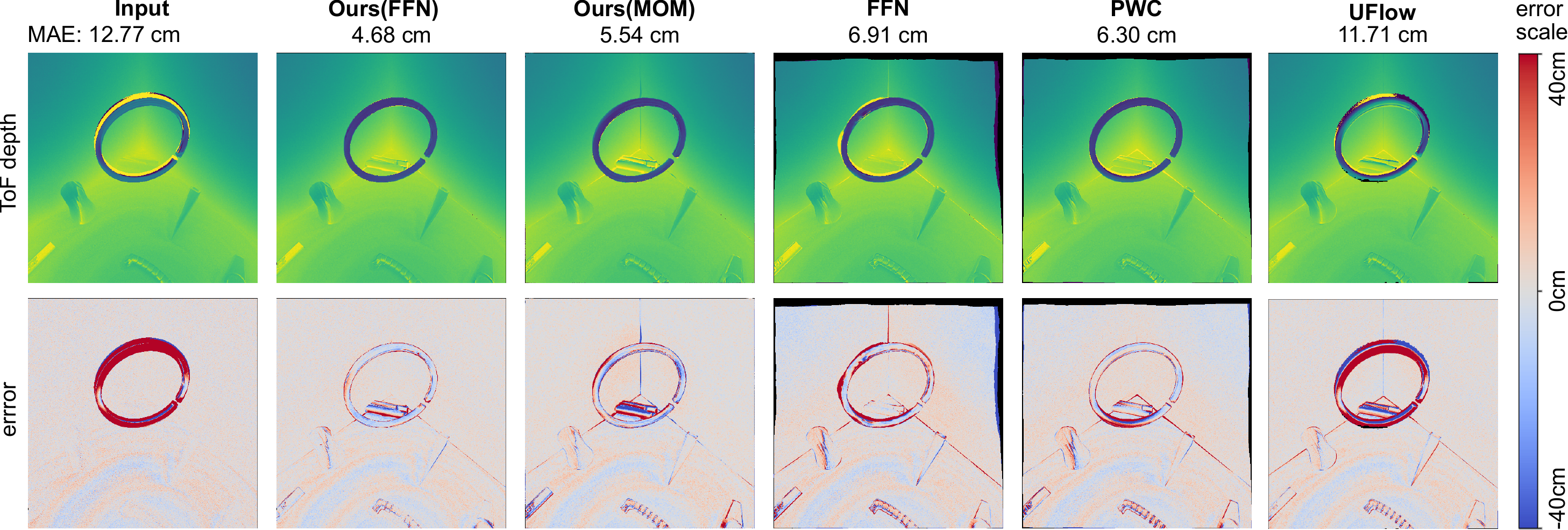}
    \caption{Motion compensation results in the single frequency single tap case. 
    Both pre-trained networks and our method resolve the motion artifacts, however our method improves performance over the pre-trained networks. Moreover, the UFlow method is not able to correct the motion artifacts.
    However, while the camera is static and only the center object is moving, all methods have some tendency to move the background, which introduces additional artifacts.
    (Empty regions after warping are shown in black.)}
    \label{fig:results_sf}
\end{figure*}

%-------------------------------------------------------------------------
\subsection{Network Architecture}

As OF backbone we investigate two networks with different architectures, the Motion Module (MOM), which was introduced by Guo \etal~\cite{guo2018tackling} for ToF motion correction, and the FFN of Kong \etal~\cite{kong2021fastflownet} which is a lightweight network with on-par performance to State-of-the-Art OF networks.
The MOM network is an encoder-decoder network based on FlowNetS~\cite{dosovitskiy2015flownet}, while the FFN integrates a latent cost volume and is based of the PWC network.
Both networks allow for fast evaluation times and low memory consumption which enables us to predict multiple flows.

While the flow prediction of the MOM network is rather straightforward, \ie it takes the set $\{m_i\}_{i=0}^{N}$ as input and predicts all flows $\{V_i\}_{i=0}^{N}$ at once, we will briefly describe how we execute the FFN in the following. Please note, that the computations of FFN are realized on a hierarchical feature pyramid, but for compact notation we neglect the hierarchy levels in the following description.

The FFN consists of the common building blocks, an image encoder $E$, a cost volume computation $C$ and a flow prediction decoder $D$.
Given the measurements $\{m_i\}_{i=0}^{N}$, we encode each measurement $m_i$ into a latent vector $F_i=E(m_i)$.
The latent vectors are then used to compute cost volumes for each pairing with the last measurement $m_{N}$, \ie $c_i = C(F_i, F_N)$ for $i=1,\dots,N-1$. 
The decoder then predicts the flows using pairs of cost volumes and latent vectors as input $V_i=D(F_i, c_i)$, the process for a single image pair is also shown on the left of Fig.~\ref{fig:losses}.
After warping the measurement $m_i$, parts of the image might remain empty, as no pixels were warped to this region, these regions are referred to as masked.

In this formulation the network only considers the two measurements $m_i, m_N$ to compute $V_i$.
Although the other measurements contain additional information about the movement, the above formulation allows to share the encoder and decoder networks for all measurements and does not increase the number of parameters.

We further apply an instance normalization to the input of the network, as also used in the ToF error correction approach of Su \etal~\cite{su2018deep}, which does not affect the depth reconstruction in Eq.~\eqref{eq:tof_depth}, as it is invariant to uniform scaling and translation of the measurements.

In case of multi-tap sensors we change the input dimension of the encoder $E$ such that it receives all measurements captured at the same time step as input.

%-------------------------------------------------------------------------
% \subsection{Multi-Tap Sensors}

%------------------------------------------------------------------------
\section{Experiments}

\begin{table}
\begin{center}
\begin{tabular}{llrrr}
\toprule
&Method & $\mathcal{L}_{photo}$ & $\mathcal{L}_{ToF}$ & mask \\
\midrule\midrule
\multirow{6}{*}{\rotatebox[origin=c]{90}{SF 1Tap}}&Input & 50.09 & 16.87 & -\\
% DeepToF & - & - & - \\
% CFN & - &  8.24 & - \\
&FFN & 54.21 & 14.63  & 12.40\%\\
&PWC & 49.16 & 13.70  & 4.12\%\\
&UFlow & {58.71} & 12.76  & 3.24\%\\
\cmidrule{2-5}
&Ours(MOM) & 34.64 & 7.64  & 0.97\%\\
&Ours(FFN) & \textbf{23.27} & \textbf{5.81} & 1.60\%\\
% \hline
% base error (noise) & & 4.22 &\\
\midrule
\multirow{8}{*}{\rotatebox[origin=c]{90}{SF 2Tap}}&Input & 34.45 & 5.93 & -\\
&FFN & 29.83 & 5.44 & 6.18\% \\
&PWC & 19.77 & 4.03 & 3.55\% \\
&UFlow & {38.22} & 4.90 & 2.07\% \\
&Lindner (FFN) & 21.01 & 4.22 & 2.35\% \\
&Lindner (PWC) & 18.11 & 3.85 & 2.12\% \\
% DeepToF & - & - & -\\
% CFN & - &  3.11 & - \\
\cmidrule{2-5}
% FFN, loss: photo  (+smooth + edge) & 19.91 & {4.20} & 99.37\\
&Ours(MOM) & 24.67 & \textbf{3.25} & 0.73\% \\
&Ours(FFN) & \textbf{17.22} & {3.66} & 0.56\%\\
% FFN, loss: ToF  (+smooth + edge + Lindner) & 16.82 & {3.59} & 99.46\\
% base error (noise) & & 1.97 &\\
% \hline
\bottomrule
\end{tabular}
\end{center}
\caption{Results for single frequency single-tap (SF 1Tap) and two tap (SF 2Tap).
The pre-trained networks, FFN and PWC, and the unsupervised UFlow method achieve only low correction rates in most cases.
The Lindner method reduces the error notably, especially when using the larger PWC as backbone, still it is outperformed by our proposed method on smaller backbones. % ($^*$ few pixels invalid due to rendering artifacts)
}
\label{tab:SF_results}
\end{table}

\begin{figure*}[t]
    \centering
    \includegraphics[width =\linewidth]{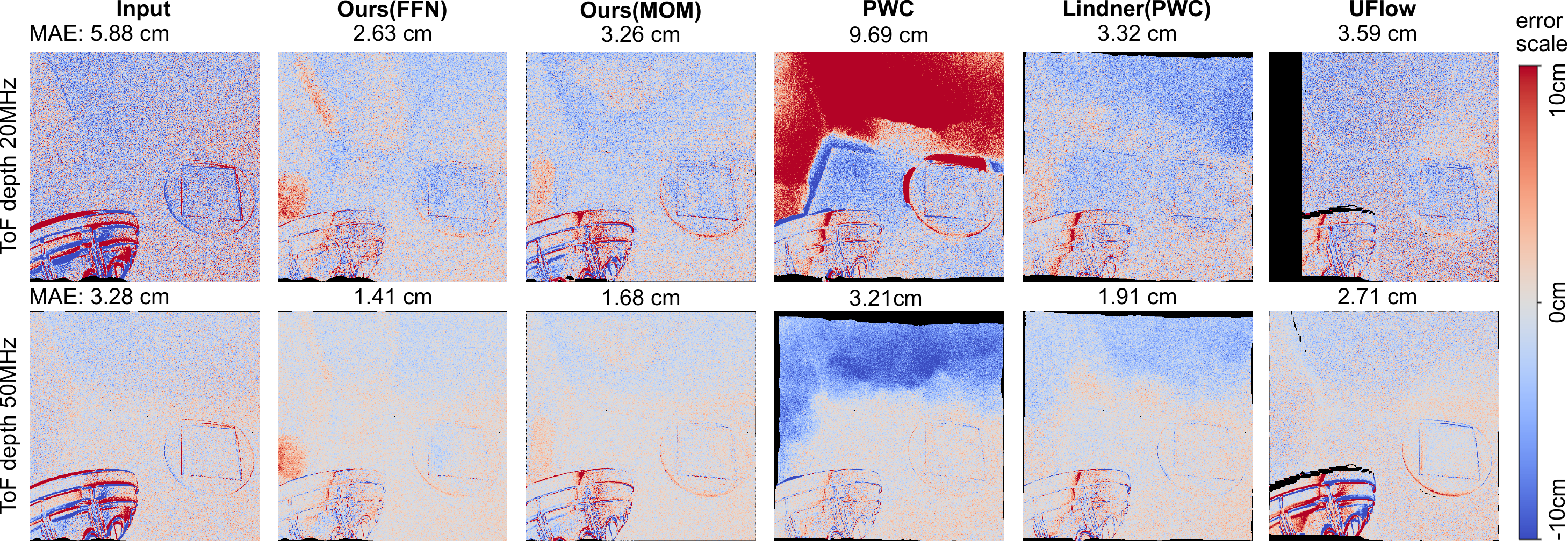}
    \caption{Motion compensation results in the multi frequency four-tap case, for a scene with moving camera.
    Our method achieves the best motion compensation, followed by Lindner's method on the more powerful backbone PWC, although Lindner's method introduces more additional errors.
    Both the pre-trained PWC and the UFlow method fail in this case.
    (Empty regions after warping are shown in black.)}
    \label{fig:results_mf}
\end{figure*}

%-------------------------------------------------------------------------
In our experiments we train instances of both FFN and MOM using the loss functions described in Sec.~\ref{sec:method}. In the case of the MOM network we do not use the similarity loss $\mathcal{L}_{sim}$, as the network does not produce latent vectors $F_i$ due to its different architectural design.
We compare against using pre-trained instances on RGB data of FFN and and also the larger PWC~\cite{sun2018pwc}, which needs $\approx$8 times the compute~\cite{kong2021fastflownet}.
In the case of multi-tap sensors we additionally compare against the Lindner method~\cite{lindner2009compensation} in combination with the pre-trained instances of FFN and PWC.
Further, we compare against the UFlow method~\cite{jonschkowski2020matters}, which is a method to train OF networks in an unsupervised fashion, and uses the PWC as backbone. 
We train the UFlow method on the same dataset as our method.

\paragraph{Dataset.}
We conduct the experiments on the CB-dataset of Schelling \etal~\cite{schelling2022radu}, as it contains raw measurements $m_i$ for three different frequencies.
It consists of 143 scenes each rendered from 50 viewpoints along a camera trajectory, which allows to simulate real movements that change the point of view.
As the CB-Dataset only incorporates static scene geometries we generated 14 additional scenes with moving objects using the same data simulation pipeline, to increase the variation of movements in the dataset.
We divide the dataset using the original training, validation and testing split, and further divide the additional scenes into 10 training scenes, and 2 each for testing and validation, whereby we use the 20MHz measurements.

%-------------------------------------------------------------------------
\subsection{Single Frequency Motion Compensation}
For the single frequency experiment we also use the 20MHz measurements of the datasets.
In the case of single-tap we take the four measurements from four subsequent time steps, in the case two-tap we take the pairs $(m_0, m_2)$ and $(m_1, m_3)$ from two times steps.
We measure $\mathcal{L}_{ToF}$, the photometric loss $\mathcal{L}_{photo}$ and the percentage of masked pixels after warping, and report results on the test set in Tab.~\ref{tab:SF_results}.
We find that the networks trained with our method achieve better results than the pre-trained OF networks and the UFlow method. Results for the single tap case can be seen in Fig.~\ref{fig:results_sf}.
The results of Lindner's method come close to our method, but only when using the larger backbone network PWC.
On the same backbone FFN the gap in performance is larger.
Additionally, in the simple setting of two-taps, and thus also two time steps, the simple MOM backbone results in better performance than the more complex FFN backbone, both trained with our method.

Further, we observe that the UFlow method increases the photometric loss, which we attribute to the fact that the method aims to minimize the photometric loss between the images of different modalities.
% As a result the network might have a tendency towards trying to warp one image into another modality.
Additionally, UFlow has a tendency to mask out areas affected by motion, as is shown in Fig.~\ref{fig:results_mf}, which leads to a reduced ToF loss, without correcting the errors.

% \noindent\textbf{Two-Tap}\\
% In the two-tap setting we additionally compare to the Lindner method~\cite{lindner2009compensation} using both a pre-trained FFN and a PWC as OF algorithm, the results are reported in Table~\ref{tab:SF_2Tap}.
%-------------------------------------------------------------------------
\subsection{Multi Frequency Motion Compensation}

\begin{table}
\setlength{\tabcolsep}{8pt}
\begin{center}
\begin{tabular}{llrrr}
\toprule
&Method & $\mathcal{L}_{photo}$ & $\mathcal{L}_{ToF}$ & mask \\
\midrule\midrule
\multirow{6}{*}{\rotatebox[origin=c]{90}{MF 1Tap}}&Input & 113.73 & 19.68 & -\\
&FFN & {124.88} & {25.06} & 10.76\% \\
&PWC & 83.15 & 16.01 & 8.91\% \\
&UFlow & {136.55} & 13.86 & 7.76\% \\
% FFN, loss: photo  (+smooth) & - & -\\
% CFN & - & 6.88 \\
% 
\cmidrule{2-5}
&Ours(MOM) & \textbf{65.91} &  \textbf{11.92} & 1.43\%\\
&Ours(FFN) & 80.43 & 13.77 & 0.34\%\\
% \hline
% base error (noise) & & 6.39 \\
\midrule
\multirow{8}{*}{\rotatebox[origin=c]{90}{MF 2Tap}}&Input & 69.06 & 8.17 & - \\
&FFN & {78.33} & {9.71} & 5.90\%\\
&PWC & 49.23 & 7.51 & 4.02\%\\
&UFlow & {81.45} & 5.95 & 4.82\%\\
&Lindner (FFN) & 40.26 & 5.60 & 2.55\% \\
&Lindner (PWC) & 35.24 & 5.16 & 1.80\% \\
% MOM & 44.68 & 4.98 & 0.64\\
% CFN & - &  3.80\\
\cmidrule{2-5}
&Ours(MOM) & 44.68 & 4.98 & 0.64\%\\
&Ours(FFN) & \textbf{30.71} & \textbf{4.43} & 0.32\% \\
\midrule
% base error (noise) & & 3.85 \\
\multirow{8}{*}{\rotatebox[origin=c]{90}{MF 4Tap}}&Input & 40.42 & 5.26 & -\\
&FFN & {57.54} & {6.93} & 0.06\%\\
&PWC  & 31.09 & {5.41} & 0.06\%\\
&UFlow & {51.10} & 4.17 & 1.96\%\\
&Lindner (FFN) & 27.52 & 3.94 & 0.06\%\\
&Lindner (PWC) & \textbf{22.17} & 3.49 & 0.06\%\\
% MOM & 29.64 & 3.11 & 0.48\\
% CFN & - &  3.74\\
\cmidrule{2-5}
&Ours(MOM) & 29.64 & 3.11 & 0.48\%\\
&Ours(FFN) & 27.14 & \textbf{3.03} & 0.08\%\\
% \hline
% base error (noise) & & x \\
\bottomrule
\end{tabular}
\end{center}
\caption{Results for multi frequency single-tap (MF 1Tap), two-tap (MF 2Tap) and four-tap (MF 4Tap).
In this setting with stronger modality changes pre-trained networks fail in most cases.
Our method is again closely followed by Lindner on the larger PWC.
}
\label{tab:MF_1Tap}
\end{table}

% \begin{table}
% \begin{center}
% \end{center}
% \caption{Results for multi frequency two-tap.}
% \label{tab:MF_2Tap}
% \end{table}

% \begin{table}
% \begin{center}
% \end{center}
% \caption{Results for multi frequency four-tap.}
% \label{tab:MF_4Tap}
% \end{table}

For the multi frequency experiment we use the three frequencies 20MHz, 50MHz and 70MHz of the datasets.
In the case of single-tap, we take the twelve measurements from twelve subsequent time steps.
In the two-tap case, we take pairs $(m_0, m_2)$ and $(m_1, m_3)$ from six time steps.
Lastly, in the case of four-tap, we use three time steps, one per frequency.
The results on the test set for both $\mathcal{L}_{ToF}$ and the photometric loss $\mathcal{L}_{photo}$ are reported in Tab.~\ref{tab:MF_1Tap}, and are shown for the four-tap case in Fig.~\ref{fig:results_mf}.

The findings from the single frequency experiment can also be observed in this setting, with our approach achieving the best performance followed by Lindner's method.
Further, the FFN trained with our method, while still outperforming the other methods, achieves rather low performance in the single tap setting, which is arguably the hardest case with the highest number of time steps, and thus the largest motion, and additionally the lowest input dimensionality of only one tap, which might make it harder for the encoder $E$ to extract modality invariant features.

Additionally, the pre-trained OF networks have a tendency to fail in these settings, especially the FFN, which might come from the larger modality gap of measurements taken at different frequencies, as can also be seen in Fig.~\ref{fig:results_mf}.

\subsection{Motion Compensation and Error Correction}

\begin{table}[t]
\setlength{\tabcolsep}{9pt}
\begin{center}
\begin{tabular}{clrr}
\toprule
& Method & MAE & Rel. Error\\
%&  & [cm] & \\
\midrule\midrule
\multirow{5}{*}{\rotatebox[origin=c]{90}{SF 1Tap}} & Input & 39.49 & 100.00\%\\
% Ours(FFN) & 31.40 & 79.51\% & 7.93 & 74.46\% \\
& CFN & 19.39 & 49.10\%\\
& CFN + Ours(FFN) & \textbf{11.47} & \textbf{29.05\%}\\
& DeepToF & 16.65 & 42.17\%\\
& DeepToF + Ours(FFN) & 15.11 & 38.26\%\\
\midrule
\multirow{7}{*}{\rotatebox[origin=c]{90}{MF 2Tap}} & Input & 10.65 & 100.00\% \\
& CFN & 6.71 & 63.01\% \\
& CFN + Ours(FFN) & \textbf{5.54} & \textbf{52.02\%} \\
& E2E & 10.44 & 98.03\% \\
& E2E + Ours(FFN)  & 8.27 & 77.65\% \\
& RADU & 11.21 & 105.26\% \\
& RADU + Ours(FFN) & 8.00 & 75.12\% \\
\bottomrule
\end{tabular}
\end{center}
\caption{Results of motion, multi-path-interference and sensor noise compensation, for the single frequency single tap (SF 1Tap) and the multi frequency two-tap (MF 2Tap) case. All methods benefit from the motion correction using our method.}
\label{tab:denoising}
\end{table}

To measure the influence on downstream error compensation techniques, we train instances of ToF correction networks on the output of our model. 
For this experiment, the single frequency single tap case and the multi frequency two-tap case are considered.
We use the single frequency approaches DeepToF~\cite{marco2017deeptof} and an adapted CFN~\cite{agresti2018deep} in the single frequency case, and the multi-frequency approaches CFN, E2E~\cite{su2018deep} and RADU~\cite{schelling2022radu} in the multi frequency case. 
For comparison we also train instances of the networks without performing motion compensation, and report results on the test set in Tab.~\ref{tab:denoising}

\begin{figure}[t]
    \centering
    \includegraphics[width =\linewidth]{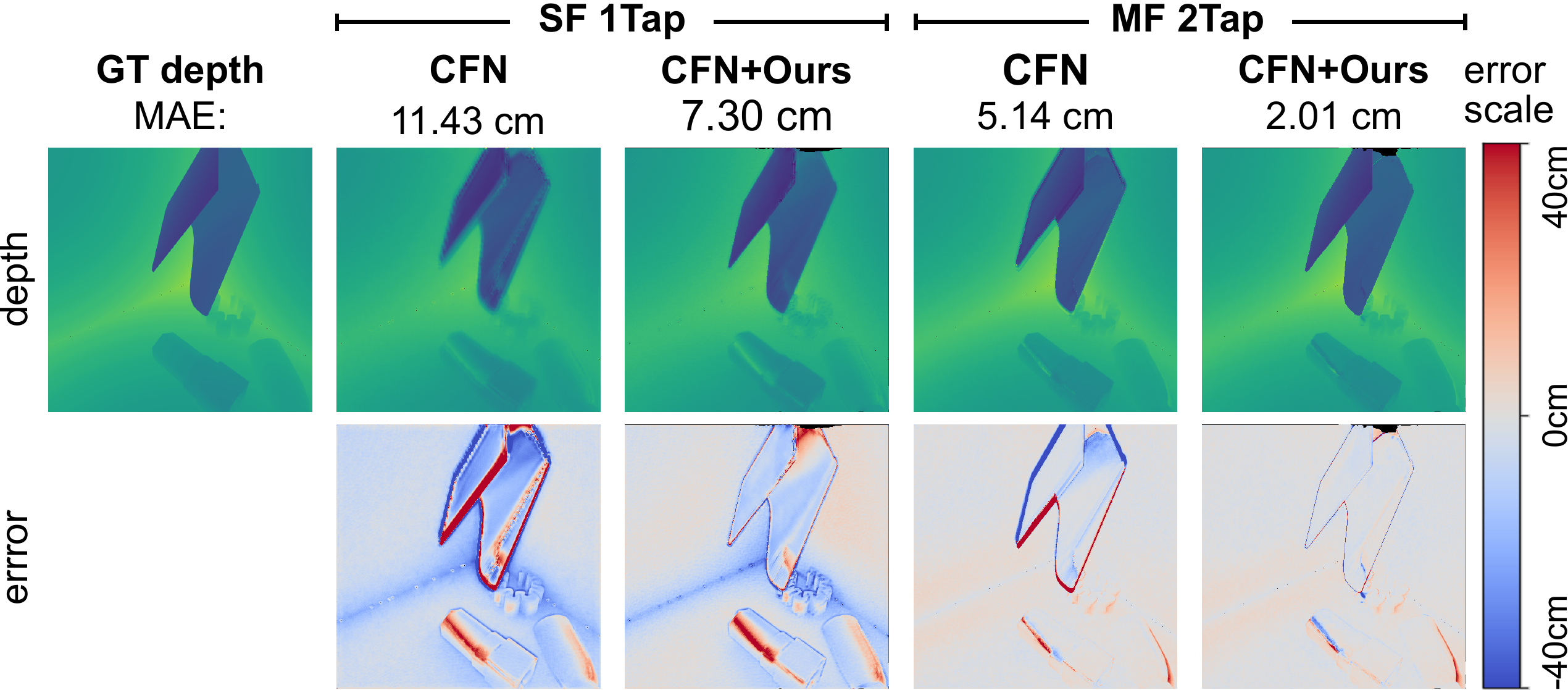}
    \caption{Results of combined motion and MPI correction using the CFN network.
    Without additional motion compensation, the motion artifacts are only partially corrected.
    In combination with method they are restricted to the object boundaries.}
    
    \label{fig:denoise}
\end{figure}

We observe, that all methods benefit from motion compensation in their input.
We further observe that the 2D networks that frame the task as denoising handle motion artifacts quite well, see Fig.~\ref{fig:denoise}, whereas the more complex approaches E2E, which formulates a generative image translation task, and RADU, which operates on 3D point clouds, struggle in this setting.
It is to be remarked, that none of the approaches were designed to correct motion artifacts.

%-------------------------------------------------------------------------
\subsection{Ablations}

This section provides ablations on the loss components.

\subsubsection{Component Ablation}

\begin{table}[b]
    \begin{center}
    \begin{tabular}{l@{\hspace{0.2em}}l@{\hspace{0.2em}}l@{\hspace{0.2em}}lrr}
    \toprule
    Method &&&& $\mathcal{L}_{photo}$ & $\mathcal{L}_{ToF}$  \\
    \midrule\midrule
    Input &&&& 70.39 & 23.71 \\
    \midrule
    $\mathcal{L}_{photo}$ + &$\mathcal{L}_{smooth}$ + &$\mathcal{L}_{edge}$ + &$\mathcal{L}_{sim}$ & 38.65 & 12.43 \\
    $\mathcal{L}_{ToF}$ &&&& 38.42 & 10.17 \\
    $\mathcal{L}_{ToF}$ + &&$\mathcal{L}_{edge}$ + &$\mathcal{L}_{sim}$ & 35.94 & 9.67 \\
    $\mathcal{L}_{ToF}$ + &$\mathcal{L}_{smooth}$ + &&$\mathcal{L}_{sim}$& 34.65 & 8.54  \\
    $\mathcal{L}_{ToF}$ + &$\mathcal{L}_{smooth}$ + &$\mathcal{L}_{edge}$ && 32.57 & {7.87}\\
    \midrule
    $\mathcal{L}_{ToF}$ + &$\mathcal{L}_{smooth}$ + &$\mathcal{L}_{edge}$ + &$\mathcal{L}_{sim}$ & \textbf{28.76} & \textbf{7.21}\\
    \bottomrule
    \end{tabular}
    \end{center}
    \caption{Ablation on the loss components in the single frequency single-tap case, using the FFN as OF backbone.}
    \label{tab:loss_abl}
\end{table}

To investigate the influence of each loss component separately, we train instances of the FFN network while disabling individual components.
Further, we replace the ToF loss $\mathcal{L}_{ToF}$ with the photometric loss $\mathcal{L}_{photo}$ and additionally train an instance using only the ToF loss as baselines.
The results on the validation set are reported in Tab.~\ref{tab:loss_abl}

From the results it can be seen that the combination of all losses achieves the best performance, and that each component reduces the loss.
Out of the regulatory losses the smoothing loss $\mathcal{L}_{smooth}$ has the highest impact, followed by the edge-aware loss $\mathcal{L}_{edge}$ and finally the latent similarity loss $\mathcal{L}_{sim}$.
Further, the ToF loss yields a large performance gain compared to the photometric loss, and even without regularizations achieves a better performance.

\subsubsection{Similarity Loss Function}

As the definition of the latent similarity loss $\mathcal{L}_{sim}$ in Eq.~\eqref{eq:L_sim} was kept general, it allows for the usage of different similarity measures $L$. 
We investigate the standard L$_1$ and L$_2$ distances, the cost function that is used in the cost volume computation and the cosine similarity
\begin{align}
    \text{L}_p: &\quad\|F_i(k,l) - F_j(k,l)\|_p, \qquad p=1,2\\
    % \text{L}2:} &\quad\|F_i(k,l) - F_j(k,l)\|_2, \\
    \text{Cost:} &\quad -F_i(k,l)\cdot F_j(k,l), \\
    \text{Cosine}: &  \quad \frac{-F_i(k,l) \cdot F_j(k,l)}{\|F_i(k,l)\|_2\|F_j(k,l)\|_2},
\end{align}
where $\cdot$ denotes the scalar product.
We consider the single frequency single tap and the multi frequency two-tap case in this ablation, and train instances of the FFN using the above similarity measures, together with all other loss components.
Further, we train an instance using no similarity loss as a baseline, and, in the case of two taps, compare to using Lindner's features as input instead of a similarity measure.
From the results, which can be seen in Tab.~\ref{tab:sim_abl}, we find that the cosine similarity achieves the best performance in both cases.
Additionally, in the multi frequency two-tap case, the cosine similarity is the only measure that improves over not using a similarity loss at all, including Lindner's method.
Consequently, both the use and the choice of the similarity measure needs careful consideration.

%------------------------------------------------------------------------
\section{Limitations}
\begin{figure}[b]
    \centering
    \includegraphics[width =\linewidth]{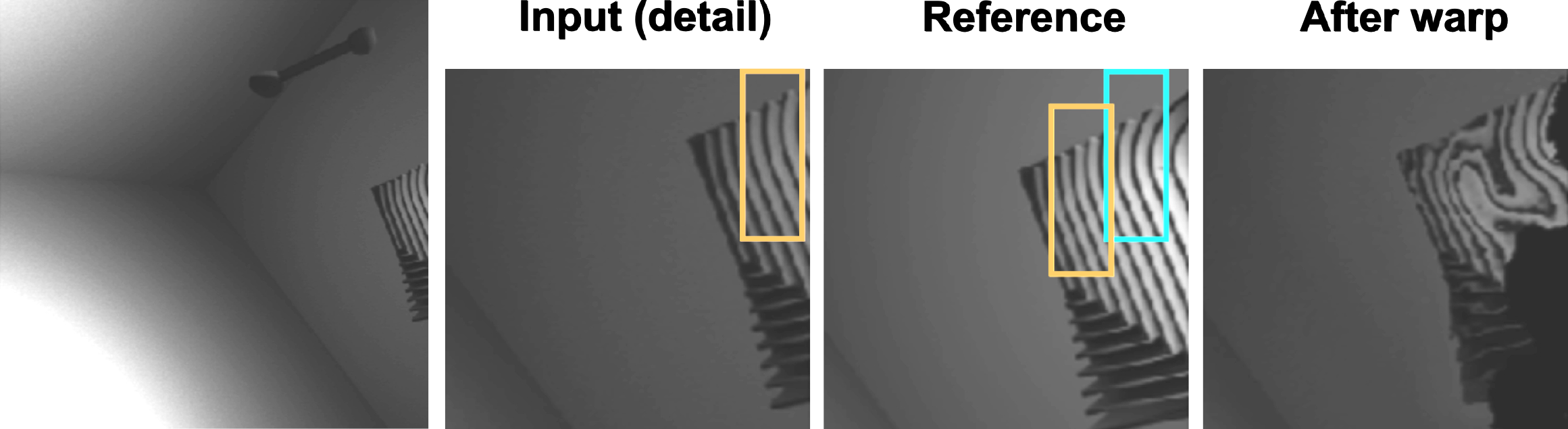}
    \caption{Example of an object, where our regularizations fail.
    The high frequency pattern prevents $\mathcal{L}_{smooth}$ from enforcing a consistent flow for the object. 
    Due to the repetitive pattern, the network matches the yellow region in the input image with the cyan region in the reference image, and the object gets distorted.
    % Although the blue region is visually similar and closer, it is not the corresponding yellow region in the reference image.
    }
    \label{fig:limitation}
\end{figure}
Although, both backbone OF networks achieve good results, we experience cases that escape our regularization losses. 
For example, the smoothing loss $\mathcal{L}_{smooth}$ ensures a continuous flow for an object, however objects are detected based on their homogeneous appearance, which can fail on high frequency details.
While the edge loss $\mathcal{L}_{edge}$ can resolve most of the cases, still sometimes wrong parts of the images are matched, especially when nearby image patches have a similar appearance, see Fig.~\ref{fig:limitation}. 
We attribute this to the fact that without access to ground truth flows, such cases present a local minima during training.

Moreover, while we demonstrated our method on the largest available iToF dataset~\cite{schelling2022radu}, this work is restricted to a synthetic setting as no real world data set containing raw iToF measurements is currently available.
Lastly, the choice of the backbone network impacts the performance in different settings, \ie MOM clearly outperforms the FFN backbone in the multi frequency single tap setting.
Additionally, as our contribution is a training algorithm, the execution time is given by the execution time of the underlying OF network, while it is almost constant in the different settings for the MOM network, it grows linearly with the number of predicted flows for the FFN.
As a consequence it would be desirable to have a OF network for iToF motion correction with a constant high performance in this multi-modality multi-frame flow prediction problem.

\begin{table}
    \setlength{\tabcolsep}{9pt}
    \begin{center}
    \begin{tabular}{lrrrr}
    \toprule
     & \multicolumn{2}{c}{SF 1Tap} & \multicolumn{2}{c}{MF 2Tap}\\
    Method & $\mathcal{L}_{photo}$ & $\mathcal{L}_{ToF}$ & $\mathcal{L}_{photo}$ & $\mathcal{L}_{ToF}$  \\
    \midrule\midrule
    Input & 70.39 & 23.71 & 93.17 & 11.98\\
    \midrule
    Lindner & - &  - & 45.59 & 7.48 \\
    None & 32.57 & 7.87 & 48.13 & 7.36 \\
    L$_1$ & 32.15 & 7.97 & 54.27 & 7.88 \\
    L$_2$ & 34.37 & 7.61 & 54.32 & 7.78 \\
    Cost & 41.88 & 10.73 & 53.98 & 7.87 \\
    Cosine & \textbf{28.76} & \textbf{7.21} & \textbf{45.49} & \textbf{6.67} \\
    % \hline
    % FlowNetS & 40.85 & 8.26 & 60.04 & 7.82 \\
    \bottomrule
    \end{tabular}
    \end{center}
    \caption{Ablation on different loss function for $\mathcal{L}_{sim}$, using FFN as backbone. On validation set.}
    \label{tab:sim_abl}
\end{table}
%------------------------------------------------------------------------
\section{Conclusion}
In this work, we presented a training method for OF networks to align iToF measurements in order to reduce the motion artifacts in the reconstructed depth images.
To this end we enable the weakly supervised training on the ToF loss $\mathcal{L}_{ToF}$ using a phase unwrapping scheme for gradient correction.
In combination with the regularizing losses $\mathcal{L}_{smooth}$ and $\mathcal{L}_{edge}$ which regulate the flow predictions, and the similarity loss $\mathcal{L}_{sim}$ to resolve the multi-modality, our method enables training without the need of ground truth flow labels.
The experiments indicate that our method is able to compensate motion artifacts for both single and multi frequency settings as well as single and multi tap sensors.
Further, our training method was demonstrated for two backbone OF networks, with different architectures, and was able to outperform existing methods.

\section{Acknowledgements}
This project was financed by the Baden-Württemberg Stiftung gGmbH.
{\small
\bibliographystyle{ieee_fullname}
\bibliography{bib}
}

\appendix

%%%%%%%%% BODY TEXT
\section{Contents}
This supplementary material provides additional information about our method in Sec.~\ref{sec:ToF_loss} and Sec.~\ref{sec:reg_loss} and further details on how the experiments were conducted in Sec.~\ref{sec:exp}.
Finally, we show additional qualitative results in Sec.~\ref{sec:qr}.

Our code, trained networks and the additional scenes to expand the CB-dataset~\cite{schelling2022radu}, containing moving objects, are available at \url{https://github.com/schellmi42/WFlowToF}.
\thispagestyle{empty}
%-------------------------------------------------------------------------
\section{Phase Unwrapping of the ToF Loss Function} \label{sec:ToF_loss}
\begin{figure}[t]
        \centering
        \includegraphics[width = \linewidth]{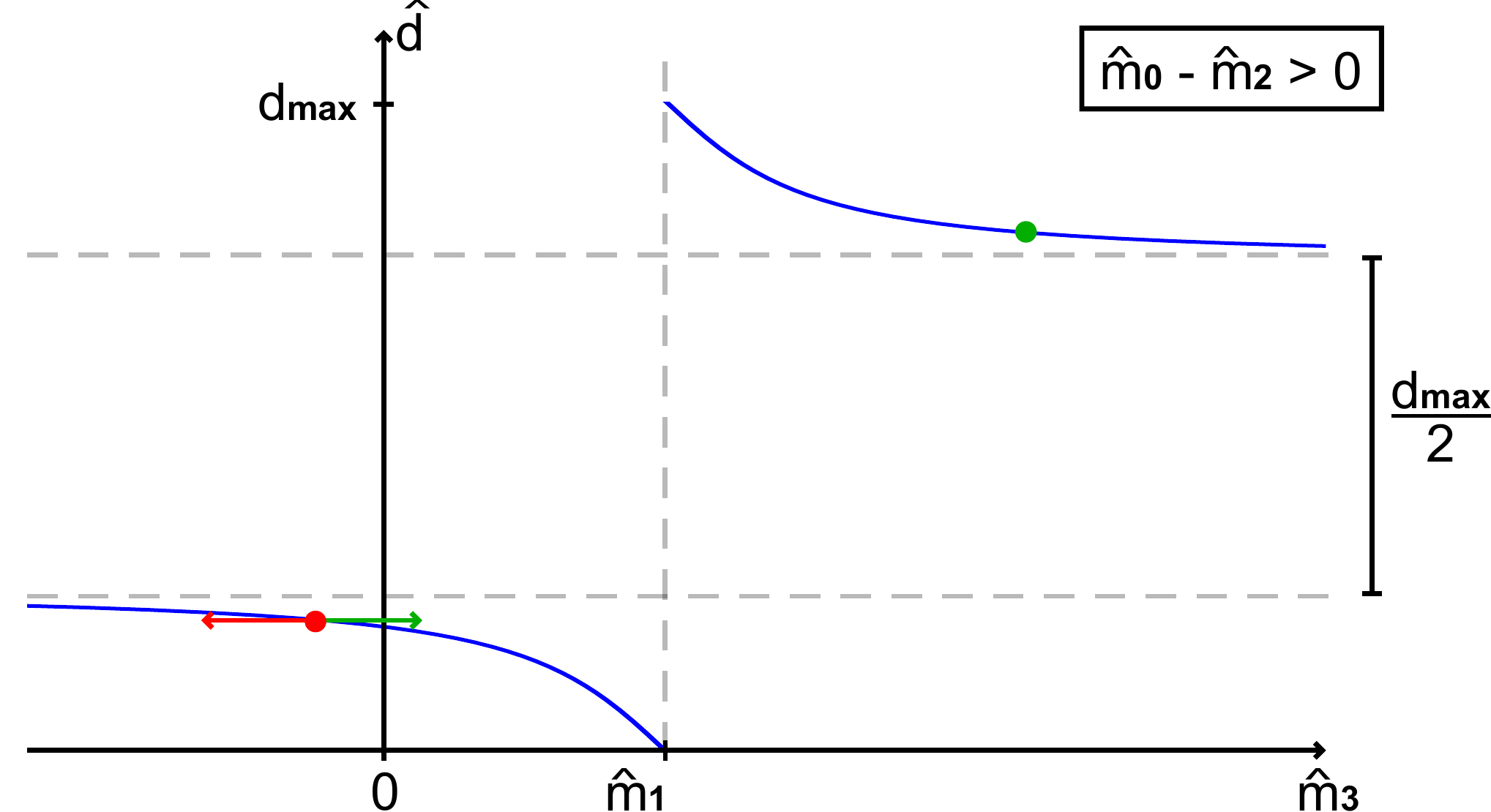}
        \caption{Reconstructed depth $\hat{d}$ dependent on the measurement $\hat{m}_0$ in the case of a positive denominator.
        The $\arctantwo$ function changes branches at $\hat{m}_3 = \hat{m}_1$, which introduces a discontinuity.
        As a result, if the prediction (red point) and the target value (green point) are on separate branches, the gradient points in the wrong direction (red arrow), and is corrected by our method (green arrow).
        The branches are separated by $d_{max}/2$, in line with Eq.~\eqref{eq:grads_PU}.}
        \label{fig:arctan_plot}
    \end{figure}

    In this section we provide more information on the phase unwrapping of the gradients of the ToF loss function $\mathcal{L}_{ToF}$, which is given through
    \begin{align}
        s &= \text{sign}(\hat{m}_0 - \hat{m}_2) \\
        \hat{d} &= \frac{c}{4\pi f}\arctan\left( \frac{\hat{m}_3 - \hat{m}_1}{\hat{m}_0 - \hat{m}_2 + s\cdot\epsilon} \right), \label{eq:tof_phase_stable} \\
        \mathcal{L}_{ToF} &= \|\hat{d} - d_{ToF}\|_1 \label{eq:L_tof},
    \end{align}
    where $\hat{m}_i$ are the iToF measurements after warping, and $\epsilon$ is a small positive constant.
    While the standard $\arctan$ function has a range limited to a semi-circle $(-\pi/2, \pi/2)$, the sign of the numerator and the denominator in Eq.~\eqref{eq:tof_phase_stable} can be used to extend the range to a full circle $(-\pi, \pi]$. This method is commonly referred to as the $\arctantwo$ function
    \begin{align}
        x &= \hat{m}_0 - \hat{m}_2, \\
        y &= \hat{m}_3 - \hat{m}_1, \\
        \hat{d} &= \frac{c}{4\pi f}\arctantwo(y, x + s\cdot\epsilon). \label{eq:tof_phase_actan2}
    \end{align}
    As a consequence, the $\arctantwo$ function has multiple branches, corresponding to the sign of its arguments, as can be seen in Fig.~\ref{fig:arctan_plot}.

    The figure also illustrates the difference of $d_{max}/2$ between the two branches in this case.
    As a result if the target (red point) is on the different branch than the current depth estimate (green point), the direction of the optimization needs to be inverted in order to move the estimate through the phase wrapping of the $\arctan2$ function and change to the correct branch.
    This is realized by our proposed gradient correction presented in the main paper
    \begin{align}
         \nabla\mathcal{L}_{ToF,PU} = 
     \left\{
     \begin{aligned}
        \nabla \mathcal{L}_{ToF}, &\qquad 0\leq \mathcal{L}_{ToF} < {d_{max} / 2},\\
        -\nabla \mathcal{L}_{ToF}, &\qquad \mathcal{L}_{ToF} \geq {d_{max} / 2}.\\
        \end{aligned}
        \right.\label{eq:grads_PU}
    \end{align}

    \begin{figure*}
        \centering
        \includegraphics[width = 0.7\linewidth]{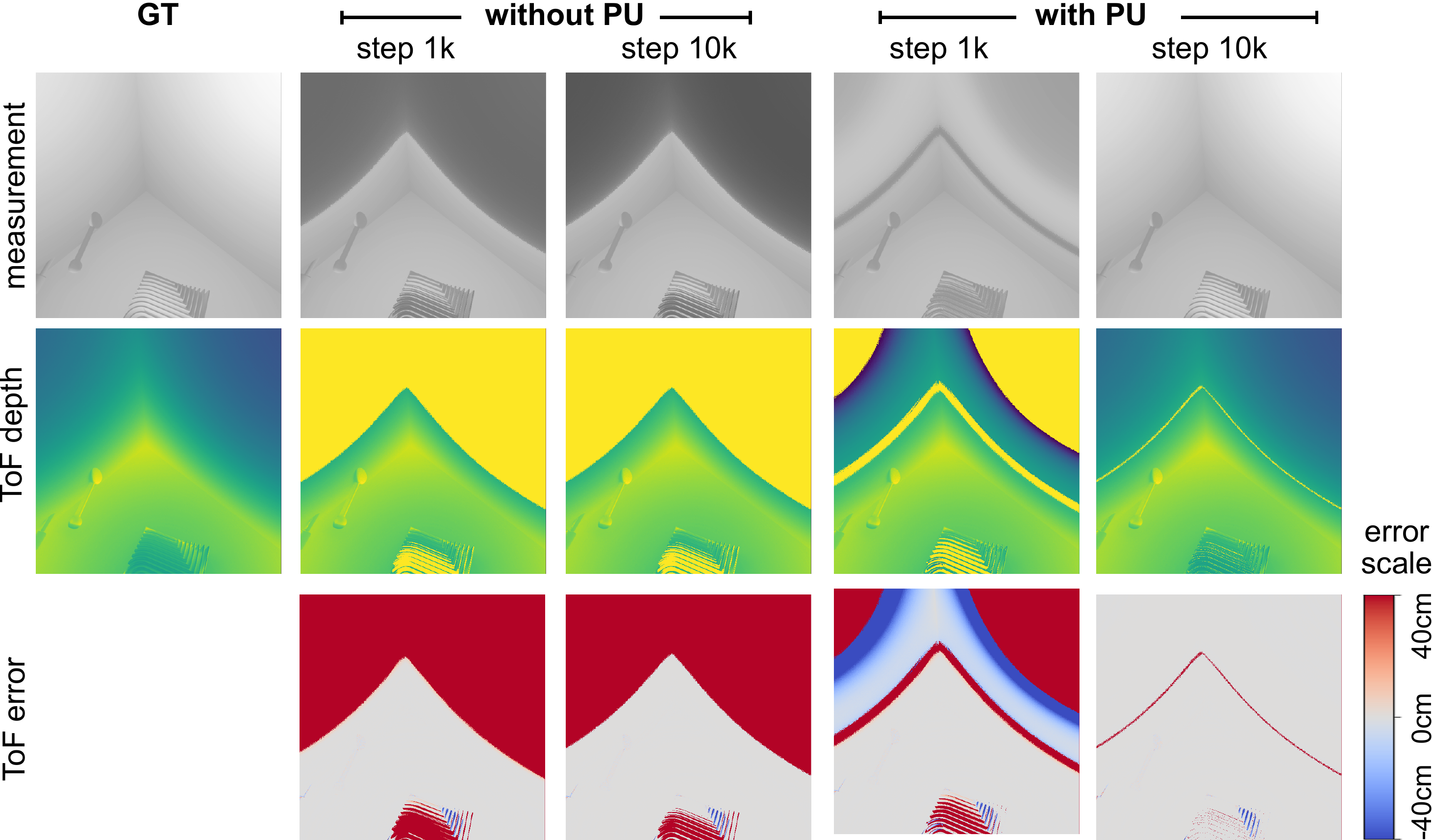}
        \caption{Reconstruction of measurement $\hat{m}_3$ by minimizing the ToF loss.
        Without phase unwrapping (PU) only partial reconstruction is possible.
        After correcting the gradients with our method, the phase wrapping is successively resolved by the optimization (right).}
        \label{fig:arctan_reconstruct}
    \end{figure*}
    
    To show the influence on the optimization we conduct a toy experiment, in which we formulate a simple reconstruction task. 
    We assume $m_0, m_1, m_2$ and $d_{ToF}$ are given and the task is to reconstruct the measurement $m_3$ by minimizing the ToF loss $\mathcal{L}_{ToF}$
    \begin{align}
        &\underset{\hat{m}_3}{\min}\ \mathcal{L}_{ToF} \\
        = &\underset{\hat{m}_3}{\min}\left\|\frac{c}{4\pi f}\arctantwo\left(\hat{m_3} - m_1, m_0 - m_2\right)-d_{ToF}\right\|_1.
    \end{align}
    We initialize $\hat{m}_0=0$ and optimize it using simple gradient descent, with and without applying our gradient correction method.
    Without gradient correction only parts initialized on the correct branch are reconstructed, wheres after gradient correction all parts can be reconstructed, as is shown in Fig.~\ref{fig:arctan_reconstruct}.
        
%------------------------------------------------------------------------
\section{Regularization Losses} \label{sec:reg_loss}
    This section provides more insights on the effect of the regularization losses.
    
    Without regularizations the problem of supervising the four measurements with a single depth is under-determined, \eg in Eq.~\eqref{eq:tof_phase_actan2} the ratio of $y / x$ is the determining factor, but not the individual values.
    While the search space is already limited when predicting optical flows, as not arbitrary values are allowed, but only values from a local neighborhood can be warped to a certain position, still regularization is necessary to further restrict the network predictions.
    Moreover, unsupervised Optical Flow (OF) networks require such regularizations in general to achieve competitive performance~\cite{jonschkowski2020matters}.

    In our work we introduced two main regularizations which measure consistency in the image space.
    The smoothing loss $\mathcal{L}_{smooth}$ measures region consistency between the predicted flow and the input image, and is applied before warping.
    The edge-aware loss $\mathcal{L}_{edge}$ measures edge consistency between the warped image and the target image, and is applied after warping.

\begin{figure}[b]
        \centering
        \includegraphics[width =\linewidth]{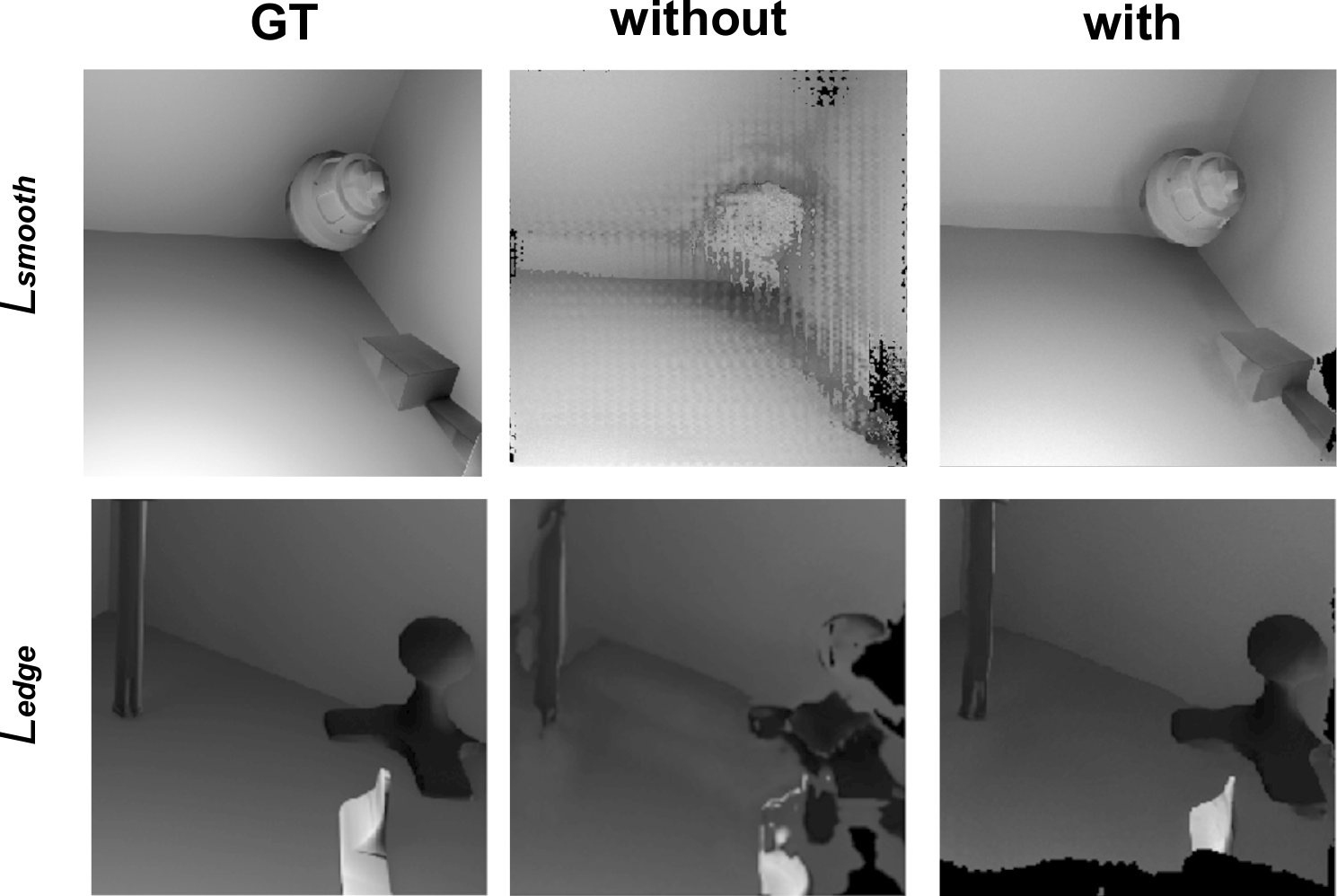}
        \caption{Effect of the regularization losses on the warped image.
        The smoothing loss $\mathcal{L}_{smooth}$ ensures that pixels that belong to a visually similar region are moved in the same direction (top row).
        With the edge aware loss $\mathcal{L}_{edge}$ the edges of the warped image align with the target image, preserving object boundaries and details. (bottom row)}
        \label{fig:reg_losses}
    \end{figure}

    The impact of these losses on the warped images can be seen in Fig.~\ref{fig:reg_losses}
            
%------------------------------------------------------------------------
\section{Experiments} \label{sec:exp}
    In this section we provide detailed information about the hyperparameters used for training the networks.

    \subsection{Implementation}
    All custom implementations were done in \verb|PyTorch 1.10.+cu102|~\cite{NEURIPS2019_9015} and \verb|Python 3.6|.
    For the OF networks FFN~\cite{kong2021fastflownet} and PWC~\cite{sun2018pwc}, and the warping operation, we use implementations provided in the PyTorch library \verb|ptlflow 0.2.5|~\cite{morimitsu2021ptlflow}.

    % Our code, trained networks and the dataset, containing moving objects, will be made publicly available online pending acceptance.

    \subsection{Motion Compensation}

    \paragraph{Our Method.}
    Both OF backbone networks FFN and MOM~\cite{guo2018tackling} are trained with the ADAM~\cite{kingma2014adam} optimizer using a learning rate scheduler which decays the learning rate by a factor $0.5$ when the ToF loss on the validation set did not decrease for 50 epochs.
    
    We augment the input data by simulating shot noise on the iToF measurements, following the noise model described by Schelling \etal~\cite{schelling2022radu}.
    Additionally, we use random image rotations by $0^\circ, 90^\circ, 180^\circ, 270^\circ$, random mirroring along the image axes, and crop random $512\times512$ image patches during training.
    
    We train the FFN network with the combination of all losses
    \begin{align}
        \mathcal{L}_{ToF} + \lambda_{smoooth} \cdot \mathcal{L}_{smooth} + \lambda_{edge}\cdot \mathcal{L}_{edge} + \lambda_{sim} \cdot \mathcal{L}_{sim}.
    \end{align}
    We compared the following values to select the hyperparameters: weights $\lambda_i$ from \{1, 1e-1, 1-e2, 1e-3, 1e-4\}, and the shift parameter $s$ in the edge-aware loss from \{1e-2, 1-e1, 0, 1e1, 1e2, 1e3\}.  
    The results reported in the main paper were achieved with $\lambda_{smooth} = 1, \lambda_{edge} = \text{1e-1},  \lambda_{sim} = \text{1e-2}, s=\text{1e2}$ in the single frequency case, and in the multi frequency case only the similarity weight was changed to $\lambda_{sim} = 1e-2$.
    In all experiments the cosine similarity was used in $\mathcal{L}_{sim}$.
    In the single frequency experiments we train with a batch size of 8, and in the multi-frequency experiment with a batch size of 4.
    The initial learning rate is set to 1e-3.

    For the MOM network we do not use the similarity loss, as the encoder decoder architecture does not have latent features for a cost volume computation, thus we set $\lambda_{sim} = 0$.
    We perform the same hyperparameter tuning as for the FFN network.
    The results in the main paper were achieved with$ \lambda_{smooth} = 1, \lambda_{edge} = 1, s=\text{1e3}$ in both the single and the multi-frequency experiments.
    We train with a batch size of 1 and an initial learning rate of 1e-5, as recommended by the authors of MOM~\cite{guo2018tackling}, 

    \paragraph{Pre-Trained Networks.}
    The pretrained networks, FFN and PWC were trained on RGB data and hence require three input channels.
    In our experiments we normalize the iToF measurements to a range of [0, 255] and repeat the the scalar image three times to match the RBG input. 
    We use weights pre-trained on the Sintel dataset~\cite{butler2012naturalistic}.
    \paragraph{UFlow.} The UFlow~\cite{jonschkowski2020matters} method is trained on the same data as our method, using the TensorFlow2 implementation provided by the authors. 
    We use the hyperparameters recommended in the documentation for custom datasets, which correspond to the settings for the Flying Chairs dataset~\cite{dosovitskiy2015flownet} in the UFlow paper.

    \paragraph{Lindner Method.}
    For the Lindner method, we match the input dimensionality of the pre-trained networks using the same scheme as above on the intensity computed with Lindner's method.

    \paragraph{Comparison to RAFT (SotA)}
    As the task involves the prediction of multiple optical flows at once, only lightweight OF networks, such as FFN, can be trained in this setting.
    More advanced architectures, such as RAFT~\cite{teed2020raft}, which is currently the State-of-the-Art (SotA) in supervised RGB OF prediction, increase the computational cost and we were unable to train it as a backbone network. 
    While a pre-trained RAFT model achieves a better performance than the simpler PWC (see Tab~\ref{tab:raft}), the inference times are much slower (see Tab.~\ref{tab:timings}). 
    Still the pre-trained RAFT model is outperformed by our method with a much smaller FFN backbone (see Tab.~\ref{tab:raft}).
    
    \begin{table}[t]
        \setlength{\tabcolsep}{4pt}
        \centering
        \begin{tabular}{lccccc}
            \hline
             & SF 1T & SF 2T & MF 1T & MF 2T & MF 4T \\
            \hline
             PWC (PT) & 13.70 & 4.03 & 16.01 & 7.51 & 5.41 \\
             RAFT (PT) & 8.48 & 4.08 & 14.72 & 6.55 & 4.63 \\
             Our (FFN) & 5.81 & 3.66 & 13.77 & 4.43 & 3.03 \\
             \hline
        \end{tabular}
        \caption{Resulting $\mathcal{L}_{ToF}$ of a pre-trained (PT) RAFT in comparison to a pre-trained PWC and our method using FFN as backbone. Results on the test set.}
        \label{tab:raft}
    \end{table}
    
    \subsection{Inference time}
    As our approach is a training algorithm, it does not affect the evaluation times of the backbone OF networks.
    As stated in the main paper, the encoder decoder architecture of MOM allows fast execution times almost independent of the number of predicted flows.
    In contrast, the runtime of networks using derivatives of the more advanced cost-volume architecture~\cite{dosovitskiy2015flownet} grows linearly corresponding to the number of predicted flows.
    Prediction times for the networks used in this work and a comparison to the SotA RAFT network are shown in Tab.~\ref{tab:timings}.
    
    \begin{table}[b]
        \setlength{\tabcolsep}{5pt}
        \renewcommand{\arraystretch}{1}
        \centering
        \begin{tabular}{lrrrrr}
            \hline
             & SF 1T & SF 2T & MF 1T & MF 2T & MF 4T \\
            \hline
             MOM & 0.002 & 0.002 & 0.002 & 0.002 & 0.002 \\
             FFN & 0.067 & 0.025 & 0.230 & 0.107 & 0.047 \\
             PWC & 0.092 & 0.054 & 0.342 & 0.154 & 0.062 \\
             RAFT & 0.342 & 0.110 & 1.275 & 0.578 & 0.228 \\
             \hline
        \end{tabular}
        \caption{Inference times in \emph{s} of the OF backbone networks, averaged over the test set, on a GTX 1080 GPU.}
        \label{tab:timings}
    \end{table}

\begin{figure*}
    \centering
    \includegraphics[width=0.9\linewidth]{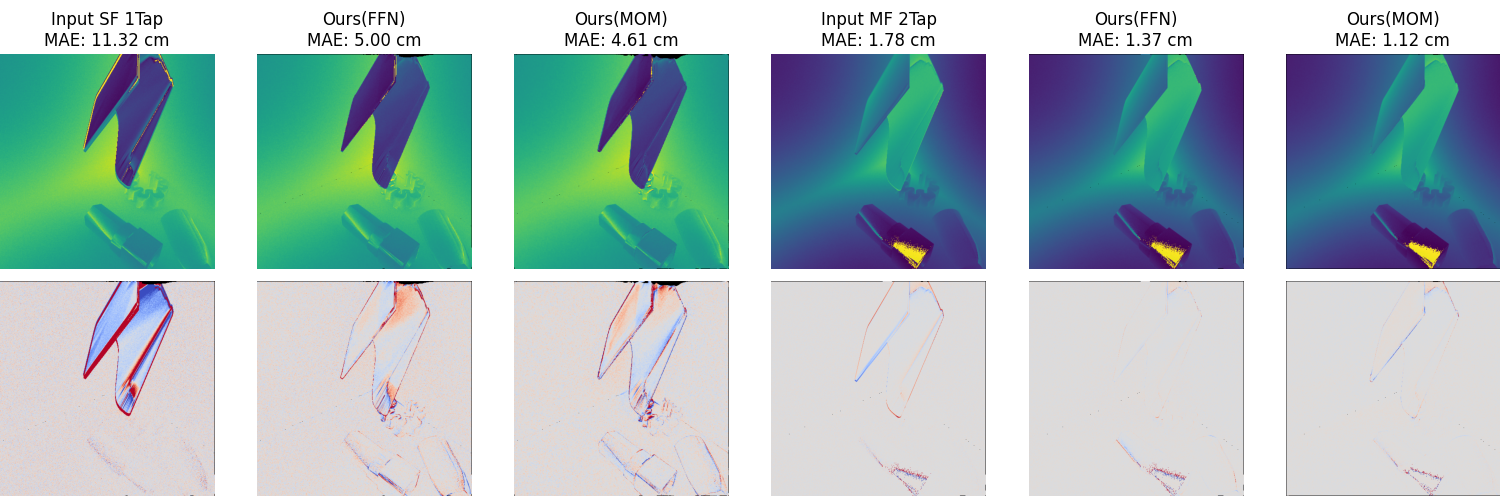}
    \includegraphics[width=0.021\linewidth]{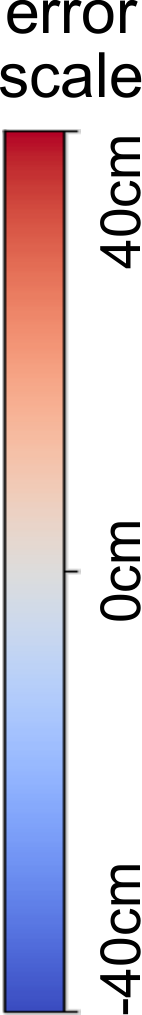}
    \includegraphics[width=0.9\linewidth]{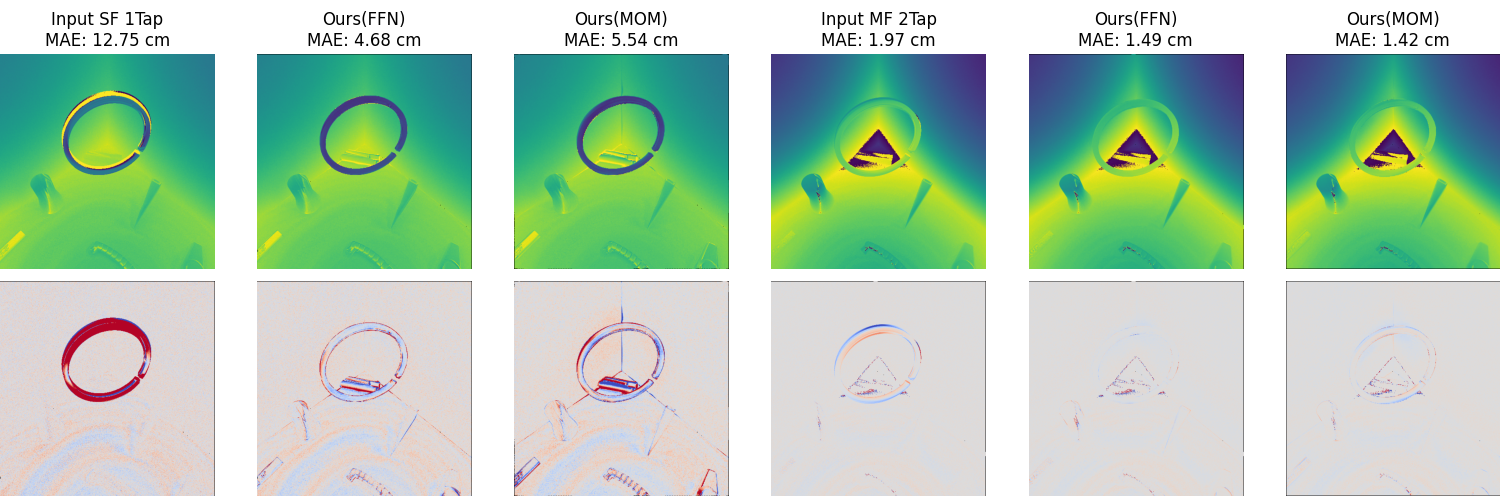}
    \includegraphics[width=0.021\linewidth]{figures_supp/results/figurebar40.pdf}
    \caption{Results of our method for single frequency single-tap (SF 1Tap, left) and multi frequency two-tap (MF 2Tap, right). The scenes contain moving objects.
    First row shows ToF depths, second row shows error maps.
    Please note that the ToF depths are not phase unwrapped.}
    \label{fig:results1}
\end{figure*}

    \subsection{Motion Compensation and Error Correction}
    We implement the CFN~\cite{agresti2018deep} in PyTorch and also adapt it to the single frequency case by reducing the input dimension to one.
    For the other approaches DeepToF~\cite{marco2017deeptof}, E2E~\cite{su2018deep} and RADU~\cite{schelling2022radu} we use the TensorFlow2 implementations by Schelling~\etal~\cite{schelling2022radu}.
    We train all networks using the respective hyperparameters reported by Schelling~\etal for their CB-Dataset.

\subsection{Ablation: Similarity Loss Function}
    We train the FFN network with the different similarity measures in the similarity loss function $\mathcal{L}_{sim}$ and optimize their weight $\lambda_{sim}$ from \{1e1, 1, 1e-1, 1e-2, 1e-3, 1e-4\} for each measure.
    When training with input generated by Lindner's method in the multi frequency two-tap case, we reduce the network input dimension to one.
    The other hyperparameters, including the weights of the other losses, are set as in the main experiments.
    The results in the main paper were achieved with the following weights:\\
    SF 1Tap:\ \ \ L1: 1e-2,\ \ L2: 1e-1,\ \ Cost: 1e-2,\ \ Cosine: 1e-3 \\
    MF 2Tap:\ \  L1: 1e-3,\ \ L2: 1e-2,\ \ Cost: 1,\ \ \ \ \ \ \ Cosine: 1e-3\\
    % \begin{tabular}{lllll}
    % SF 1Tap:&L1: 1e-2,&L2: 1e-1,& Cost: 1e-2,& Cosine: 1e-3 \\
    % MF 2Tap:&  L1: 1e-3,& L2: 1e-2,&Cost: 1,& Cosine: 1e-3\\
    % \end{tabular}

\section{Qualitative Results} \label{sec:qr}
\begin{figure*}
    \centering
    \includegraphics[width=0.9\linewidth]{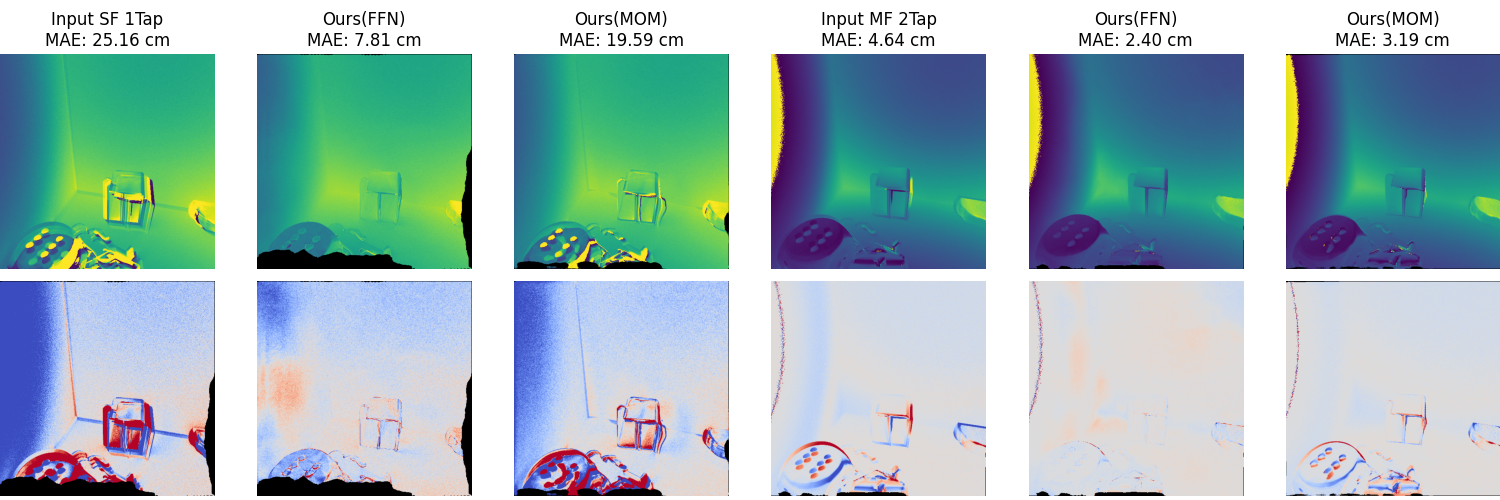}
    \includegraphics[width=0.021\linewidth]{figures_supp/results/figurebar40.pdf}
    \includegraphics[width=0.9\linewidth]{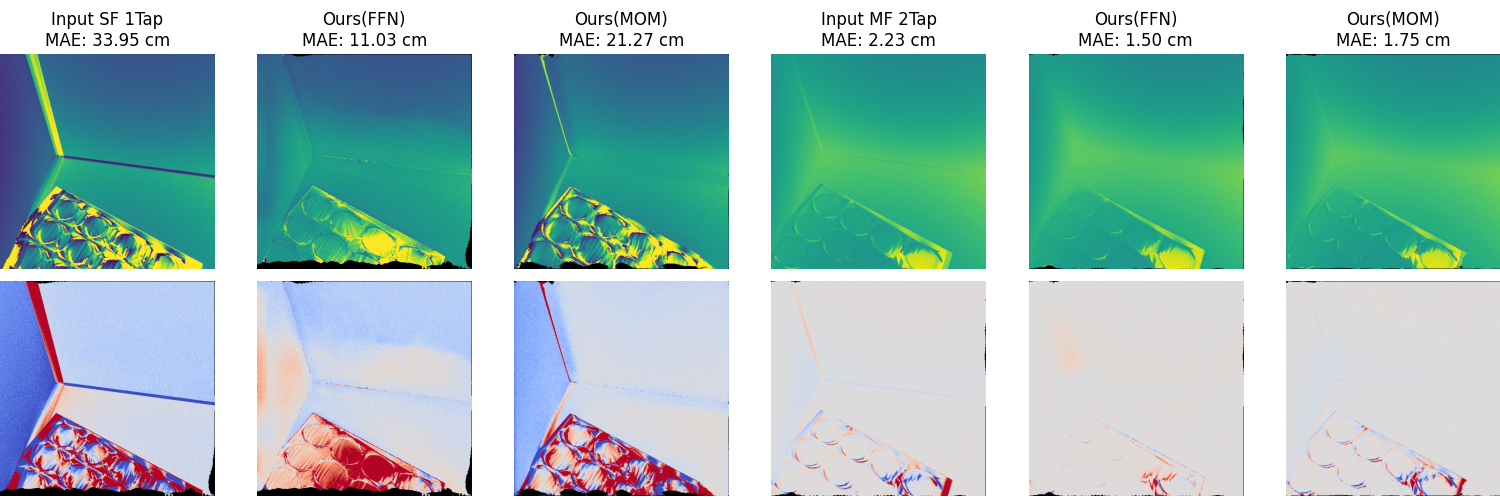}
    \includegraphics[width=0.021\linewidth]{figures_supp/results/figurebar40.pdf}
    \includegraphics[width=0.9\linewidth]{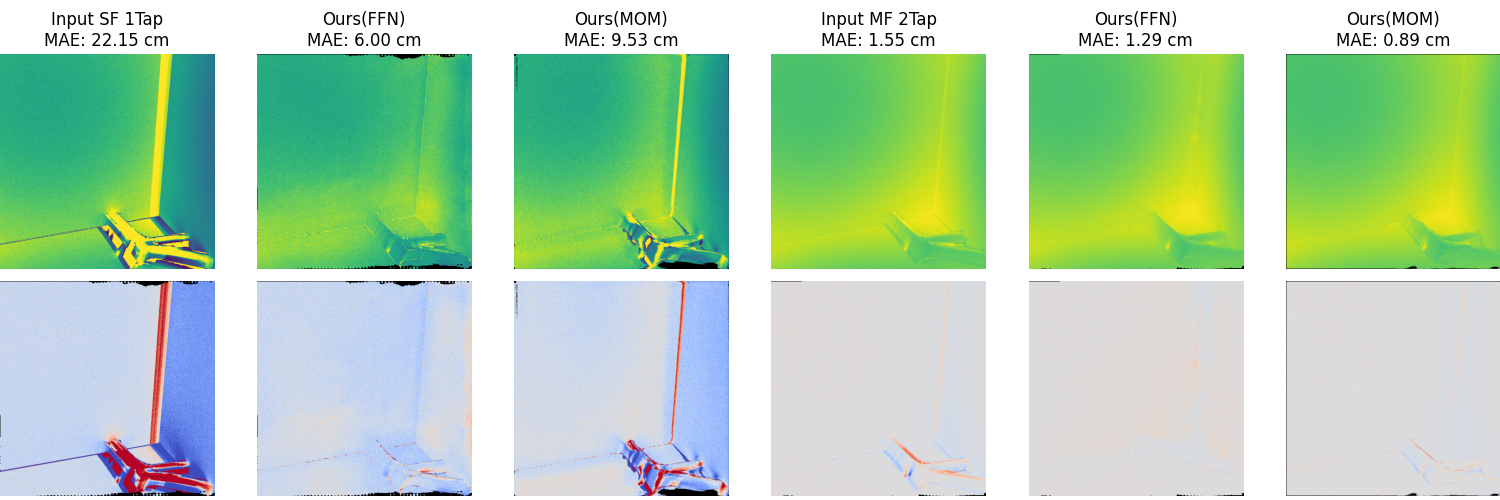}
    \includegraphics[width=0.021\linewidth]{figures_supp/results/figurebar40.pdf}
    \includegraphics[width=0.9\linewidth]{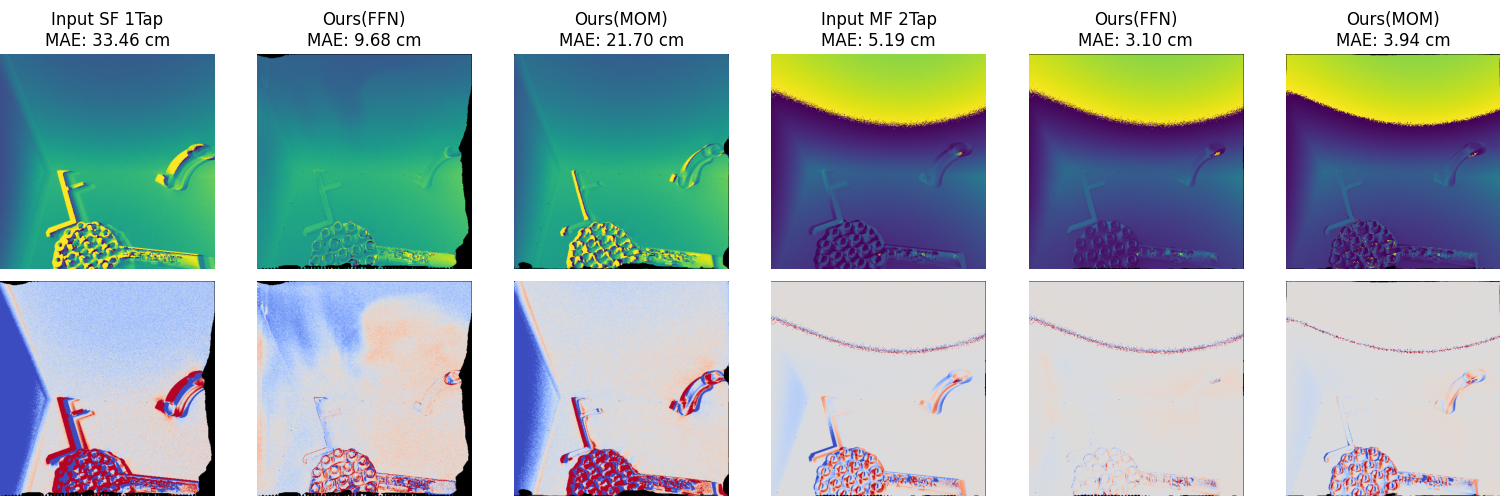}
    \includegraphics[width=0.021\linewidth]{figures_supp/results/figurebar40.pdf}
    \caption{Results of our method for single frequency single-tap (SF 1Tap, left) and multi frequency two-tap (MF 2Tap, right). 
    First row shows ToF depths, second row shows error maps.
    ToF depths are not phase unwrapped.}
    \label{fig:results2}
\end{figure*}

\begin{figure*}
    \centering
    \includegraphics[width=0.9\linewidth]{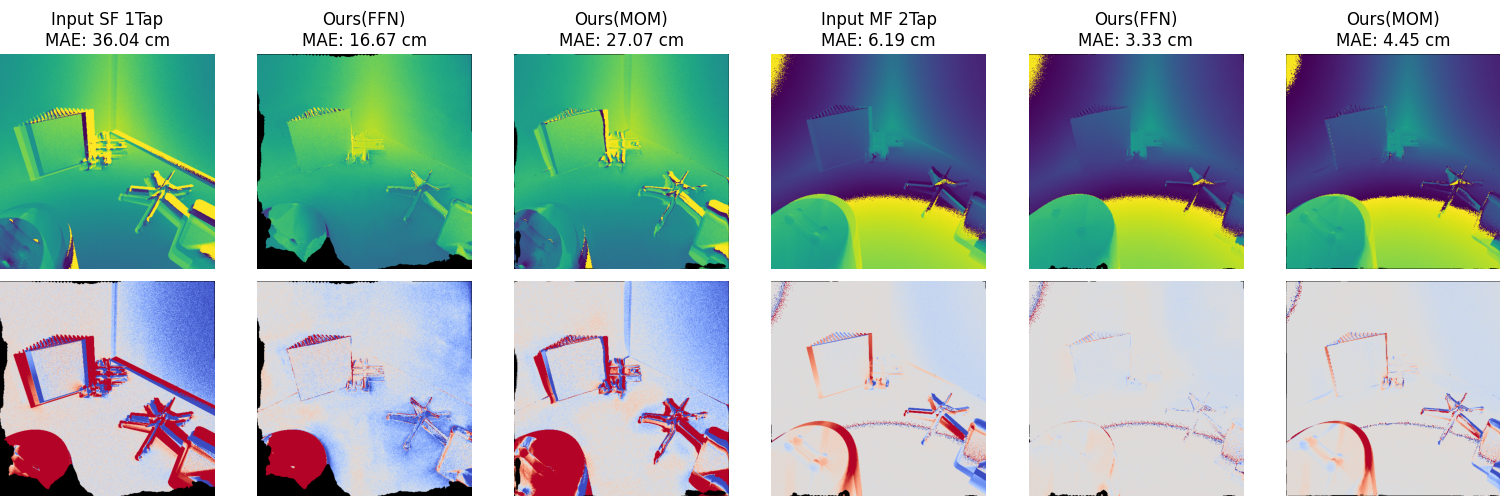}
    \includegraphics[width=0.021\linewidth]{figures_supp/results/figurebar40.pdf}
    \includegraphics[width=0.9\linewidth]{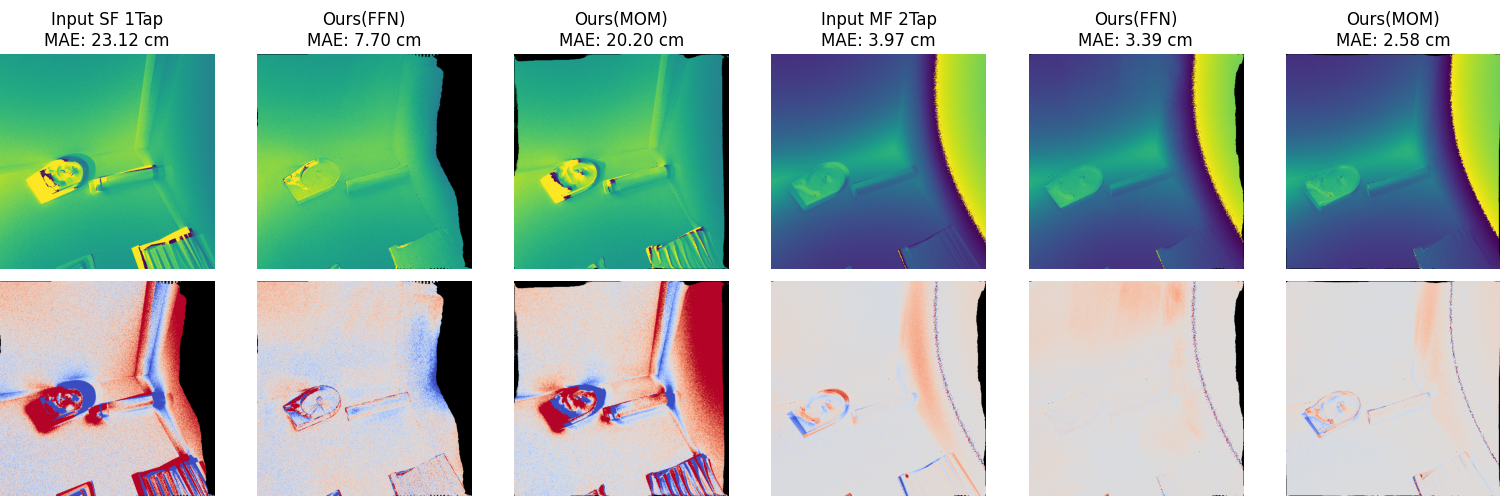}
    \includegraphics[width=0.021\linewidth]{figures_supp/results/figurebar40.pdf}
    \includegraphics[width=0.9\linewidth]{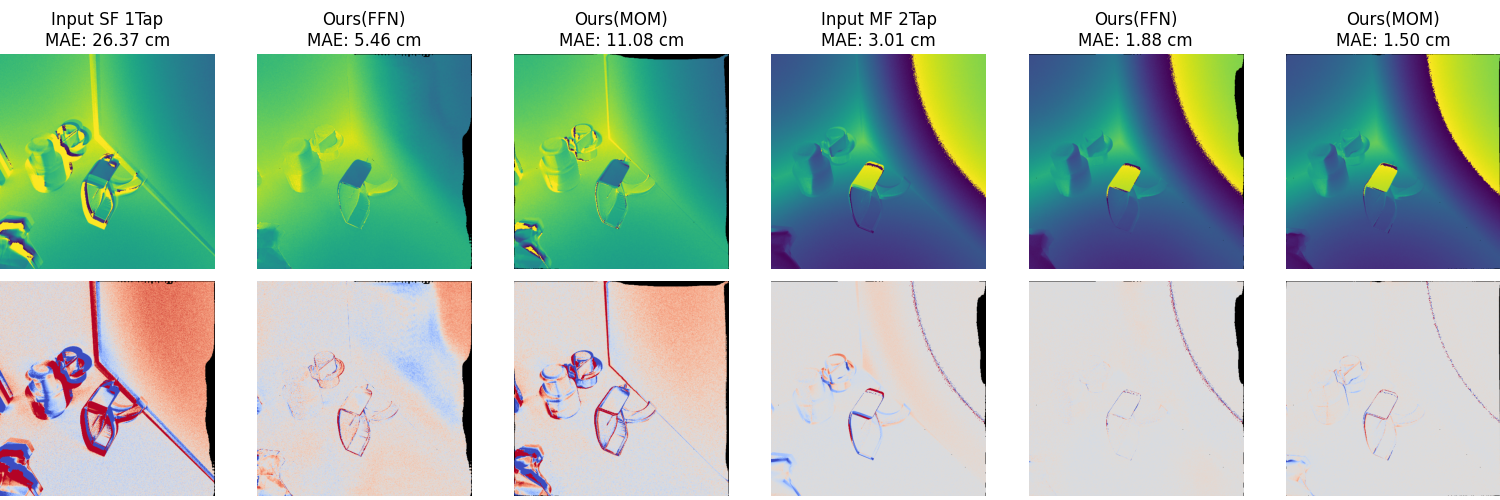}
    \includegraphics[width=0.021\linewidth]{figures_supp/results/figurebar40.pdf}
    \includegraphics[width=0.9\linewidth]{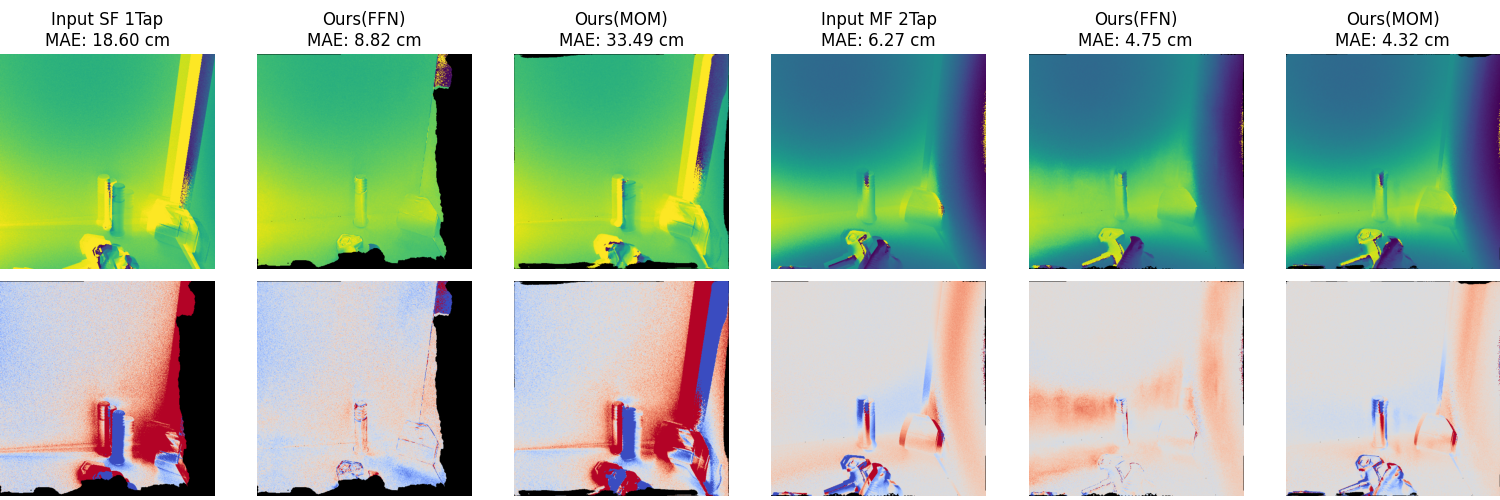}
    \includegraphics[width=0.021\linewidth]{figures_supp/results/figurebar40.pdf}
    \caption{Results of our method for single frequency single-tap (SF 1Tap, left) and multi frequency two-tap (MF 2Tap, right). 
    First row shows ToF depths, second row shows error maps.
    Please note that the ToF depths are not phase unwrapped.}
    \label{fig:results3}
\end{figure*}

\begin{figure*}
    \centering
    \includegraphics[width=0.9\linewidth]{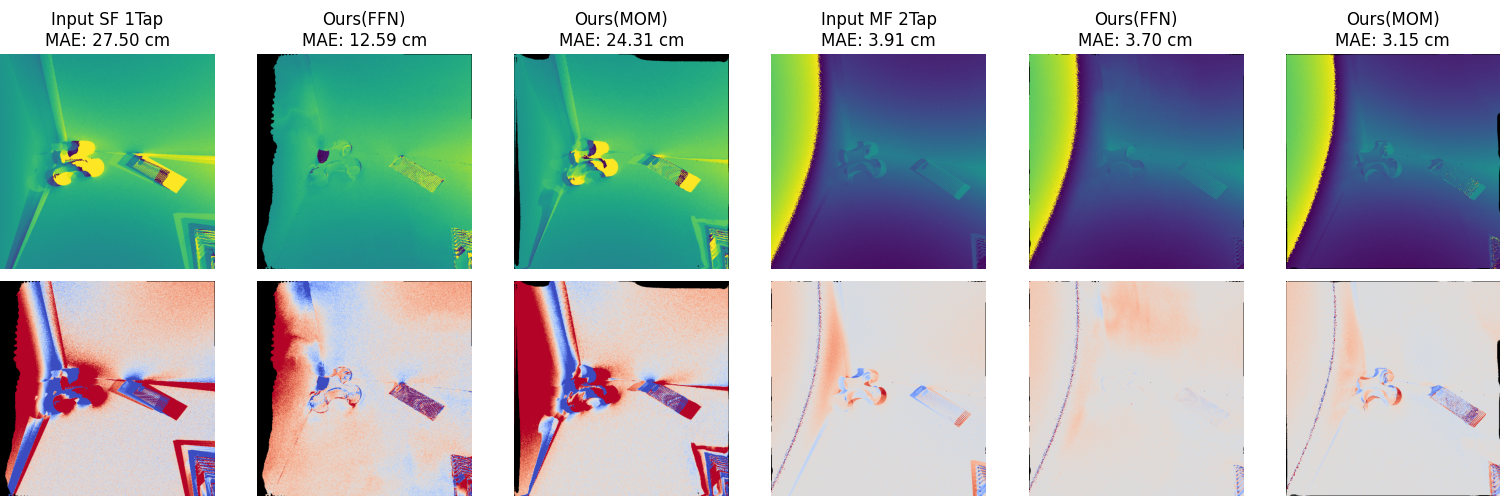}
    \includegraphics[width=0.021\linewidth]{figures_supp/results/figurebar40.pdf}
    \includegraphics[width=0.9\linewidth]{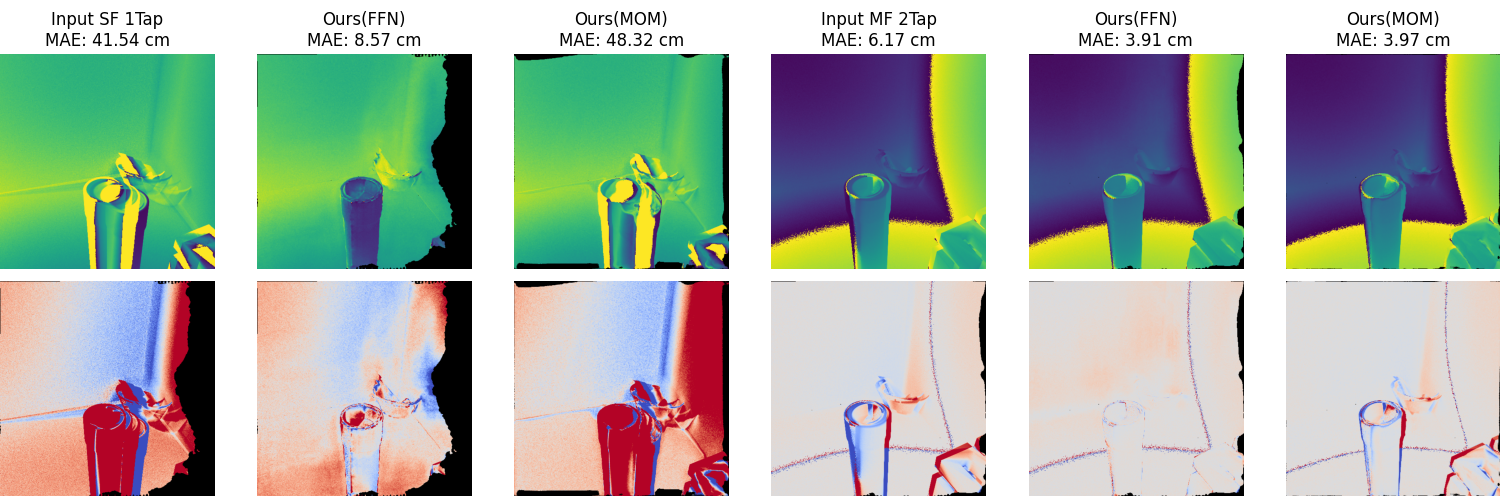}
    \includegraphics[width=0.021\linewidth]{figures_supp/results/figurebar40.pdf}
    \includegraphics[width=0.9\linewidth]{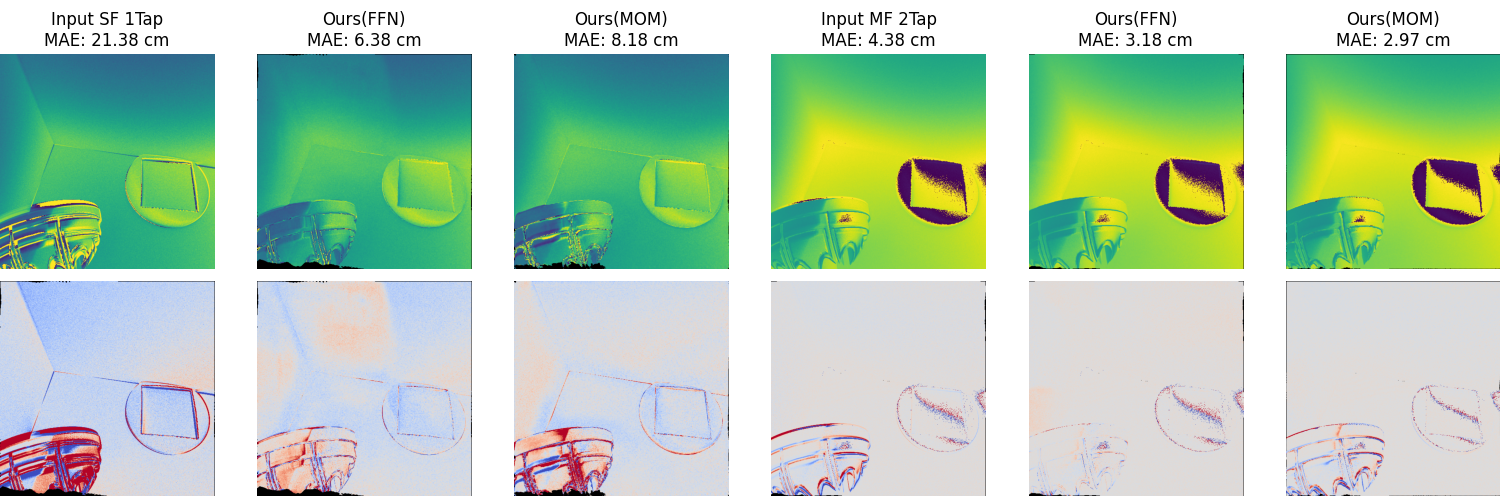}
    \includegraphics[width=0.021\linewidth]{figures_supp/results/figurebar40.pdf}
    \includegraphics[width=0.9\linewidth]{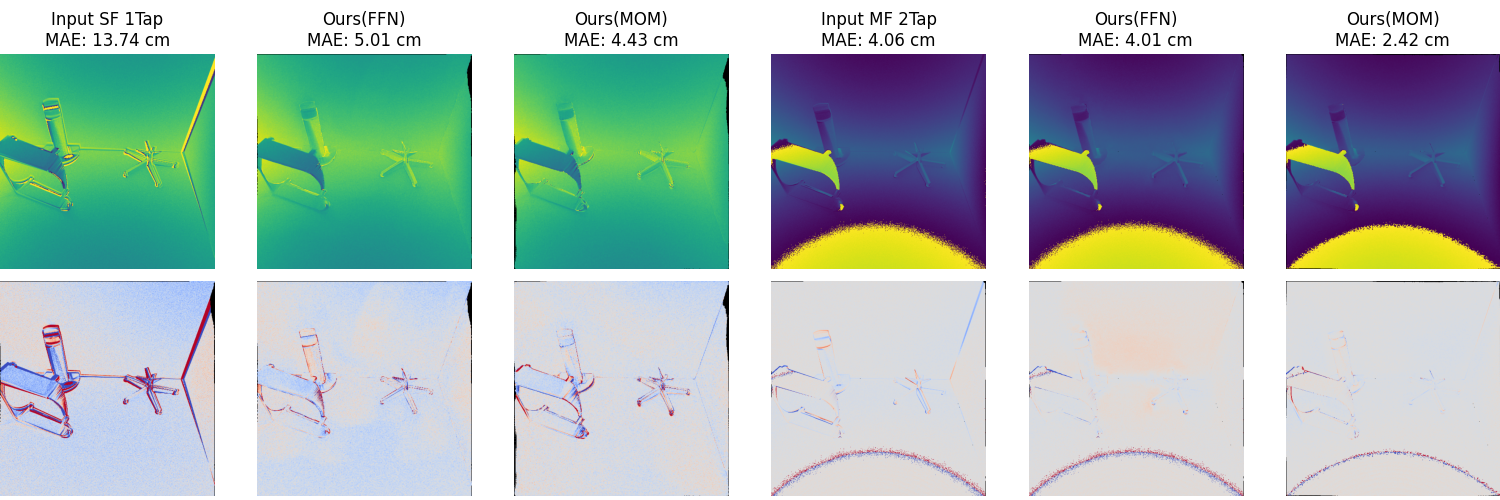}
    \includegraphics[width=0.021\linewidth]{figures_supp/results/figurebar40.pdf}
    \caption{Results of our method for single frequency single-tap (SF 1Tap, left) and multi frequency two-tap (MF 2Tap). 
    First row shows ToF depths, second row shows error maps.
    Please note that the ToF depths are not phase unwrapped.}
    \label{fig:results4}
\end{figure*}

    Results of our method using the FFN and the MOM network can be seen in Fig.~\ref{fig:results1},~\ref{fig:results2},~\ref{fig:results3} and~\ref{fig:results4}.
    The figures show one frame per scene from the test set.
    To cover both single and multi frequency and single and multi-tap the two cases single frequency single tap and multi frequency two tap are shown.

    % The FFN compensates the motion artifacts in all scenes, whereas the MOM backbone fails in some cases with relative large motions.

\begin{figure*}
    \centering
    \includegraphics[width=0.95\linewidth]{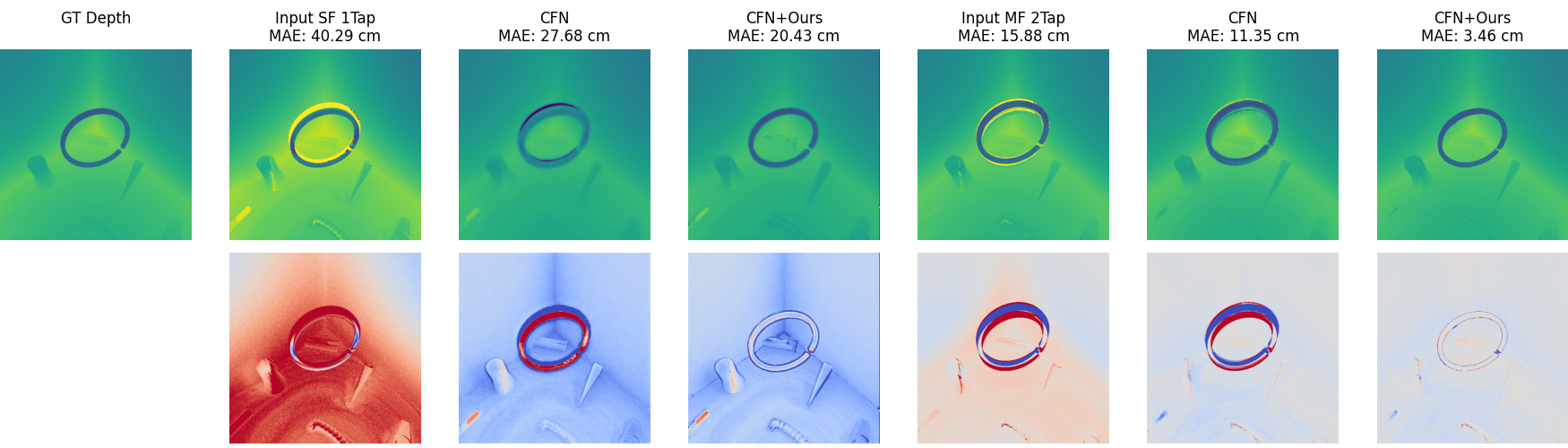}
    \includegraphics[width=0.019\linewidth]{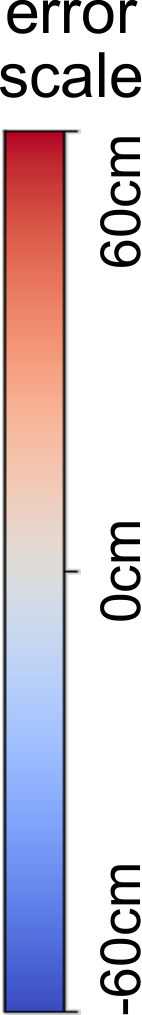}
    \includegraphics[width=0.95\linewidth]{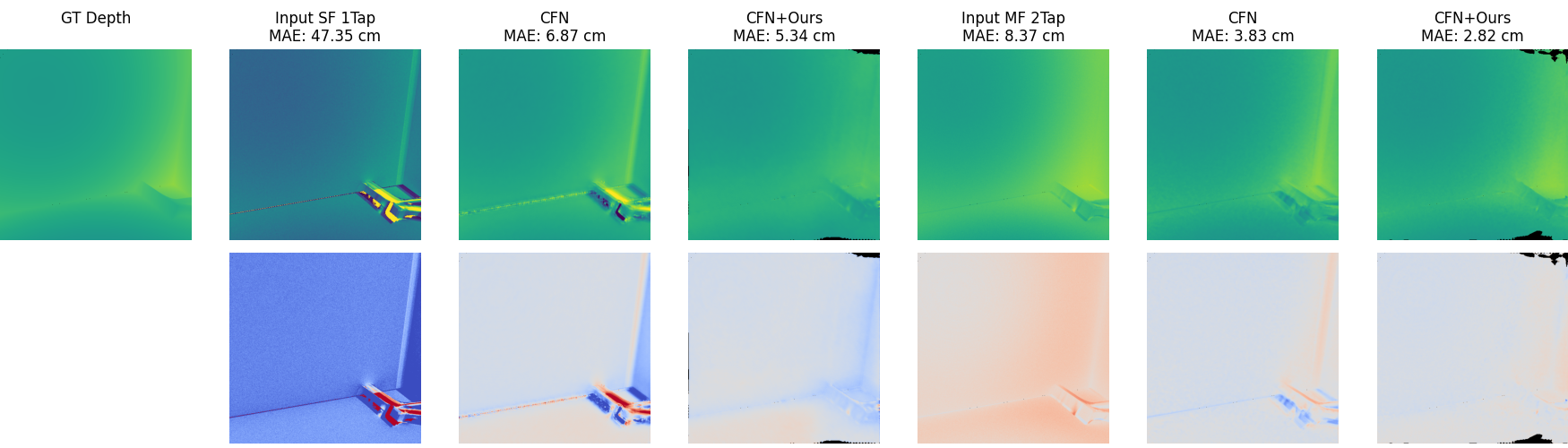}
    \includegraphics[width=0.019\linewidth]{figures_supp/denoising/figurebar60.pdf}
    \includegraphics[width=0.95\linewidth]{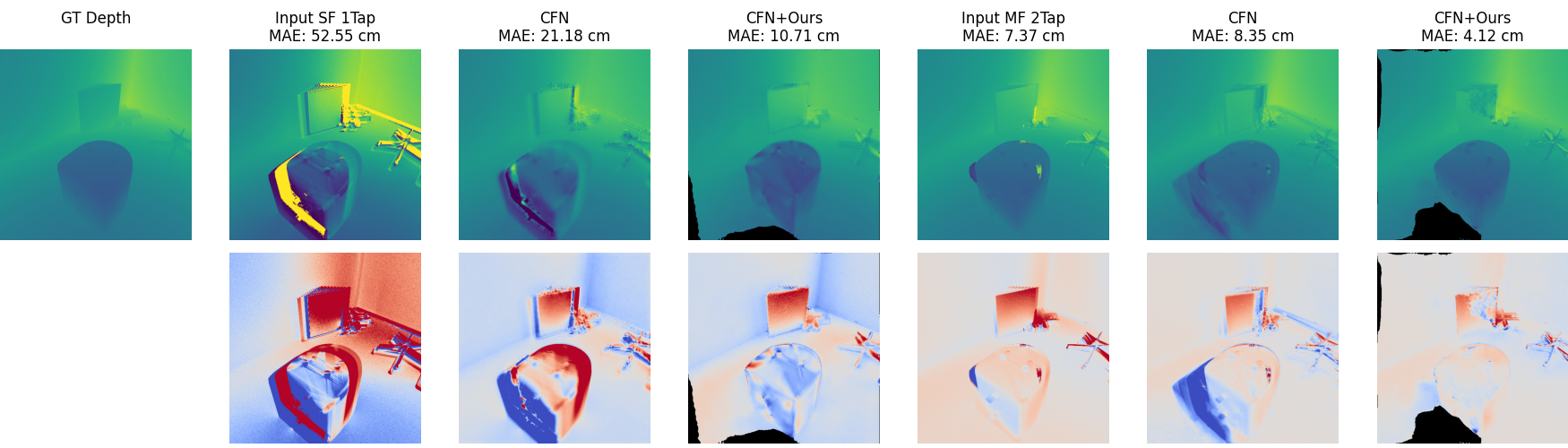}
    \includegraphics[width=0.019\linewidth]{figures_supp/denoising/figurebar60.pdf}
    \includegraphics[width=0.95\linewidth]{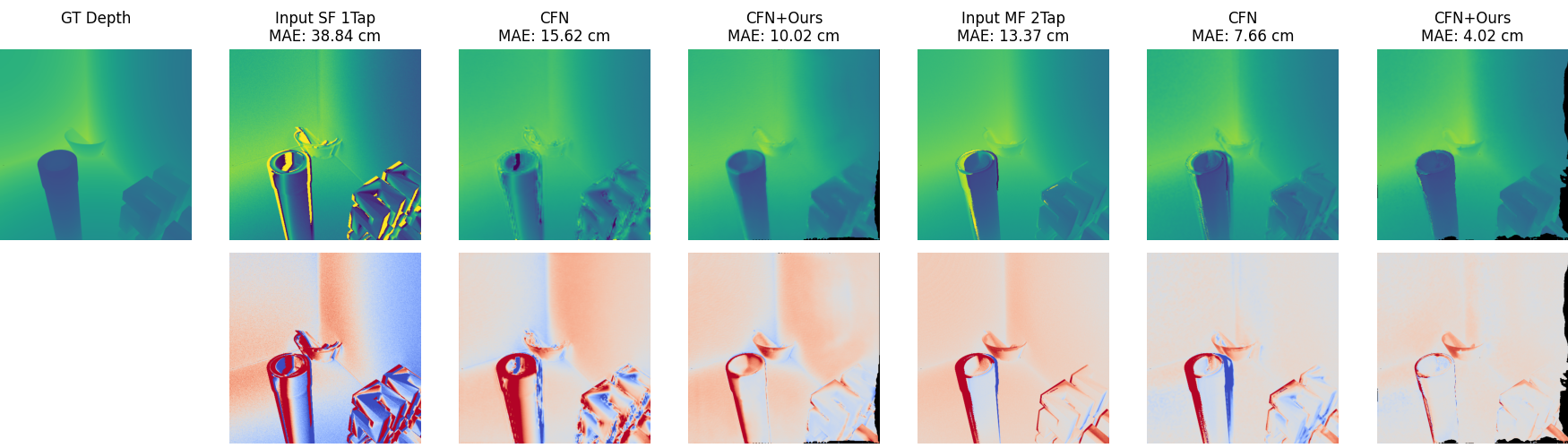}
    \includegraphics[width=0.019\linewidth]{figures_supp/denoising/figurebar60.pdf}
    \caption{Results of combined motion and MPI correction using the CFN network. 
    First row shows depths, second row shows error maps.
    Input in the MF 2Tap case is shown at highest frequency.
    First scene contains a moving object. 
    Please note that CFN receives input from all frequencies, which can result in additional motion artifacts in the prediction compared to the high frequency input ToF depth.}
    \label{fig:results_combined}
\end{figure*}

    In Fig.~\ref{fig:results_combined} we show additional results for the combined correction of motion artifacts and Multi-Path-Interference (MPI) using the CFN as error correction network.

% \begin{figure*}
%     \centering
%     \includegraphics[width=0.9\linewidth]{figures_supp/results/013.png}
%     \includegraphics[width=0.021\linewidth]{figures_supp/results/figurebar40.pdf}
%     \includegraphics[width=0.9\linewidth]{figures_supp/results/014.png}
%     \includegraphics[width=0.021\linewidth]{figures_supp/results/figurebar40.pdf}
%     \caption{Results of our method for single frequency single-tap (SF 1Tap, left) and multi frequency two-tap (MF 2Tap, right). 
%     First row shows ToF depths, second row shows error maps.
%     ToF depths are not phase unwrapped.}
%     \label{fig:results4}
% \end{figure*}

\end{document}